\documentclass[review]{elsarticle}

\usepackage{lineno,hyperref}
\modulolinenumbers[5]

\journal{Journal of \LaTeX\ Templates}

\let\oldFootnote\footnote
\newcommand\nextToken\relax

\renewcommand\footnote[1]{%
    \oldFootnote{#1}\futurelet\nextToken\isFootnote}

\newcommand\isFootnote{%
    \ifx\footnote\nextToken\textsuperscript{,}\fi}
    
\usepackage{subcaption}
\usepackage{cite}
\usepackage{multirow}
\usepackage{adjustbox}
\usepackage[table,xcdraw,pdftex,dvipsnames]{xcolor}
\usepackage{xurl}
\usepackage{xargs} 
\usepackage{amsmath,amssymb,bm}
\usepackage{mathtools}
\usepackage{colortbl}
\usepackage{soul}
\usepackage{tabularx}
\usepackage{bigfoot}
\DeclareNewFootnote[para]{default}
\usepackage{arydshln}
\usepackage{color} 
\definecolor{grey}{rgb}{0.7, 0.7, 0.7}

\usepackage[colorinlistoftodos,prependcaption,textsize=tiny]{todonotes}

\newcommandx{\todoresolve}[2][1=]{\todo[linecolor=Plum,backgroundcolor=Plum!25,bordercolor=Plum,#1]{#2}}
\newcommandx{\todoresolved}[2][1=]{\todo[linecolor=Blue,backgroundcolor=Blue!25,bordercolor=Blue,#1]{#2}}
\newcommandx{\todotext}[2][1=]{\todo[linecolor=OliveGreen,backgroundcolor=OliveGreen!25,bordercolor=OliveGreen,#1]{#2}}
\newcommandx{\todoadd}[2][1=]{\todo[linecolor=Red,backgroundcolor=Red!25,bordercolor=Red,#1]{#2}}

\begin{document}
\begin{frontmatter}

    \title{A Commonsense-Infused Language-Agnostic Learning Framework for Enhancing Prediction of Political Polarity in Multilingual News Headlines}

    \author[1,2]{Swati Swati\corref{cor1}}
            \ead{swati@ijs.si}
            \cortext[cor1]{Corresponding author.}
        \author[1]{Adrian Mladenić Grobelnik}
            \ead{adrian.m.grobelnik@ijs.si}
        \author[1,2]{Dunja Mladenić}
            \ead{dunja.mladenic@ijs.si}
        \author[1]{Marko Grobelnik}
            \ead{marko.grobelnik@ijs.si}
            
        \address[1]{Jožef Stefan Institute, Jamova cesta 39, 1000 Ljubljana, Slovenia}
        \address[2]{Jožef Stefan International Postgraduate School, Jamova cesta 39, 1000 Ljubljana, Slovenia}

    \begin{abstract}
        
        Predicting the political polarity of news headlines is a challenging task that becomes even more challenging in a multilingual setting with low-resource languages. To deal with this, we propose to utilise the Inferential Commonsense Knowledge via a Translate-Retrieve-Translate strategy to introduce a learning framework. To begin with, we use the method of translation and retrieval to acquire the inferential knowledge in the target language. We then employ an attention mechanism to emphasise important inferences. We finally integrate the attended inferences into a multilingual pre-trained language model for the task of bias prediction. To evaluate the effectiveness of our framework, we present a dataset of over 62.6K multilingual news headlines in five European languages annotated with their respective political polarities. We evaluate several state-of-the-art multilingual pre-trained language models since their performance tends to vary across languages (low/high resource). Evaluation results demonstrate that our proposed framework is effective regardless of the models employed. Overall, the best performing model trained with only headlines show 0.90 accuracy and F1, and 0.83 jaccard score. With attended knowledge in our framework, the same model show an increase in 2.2\% accuracy and  F1, and 3.6\% jaccard score. Extending our experiments to individual languages reveals that the models we analyze for Slovenian perform significantly worse than other languages in our dataset. To investigate this, we assess the effect of translation quality on prediction performance. It indicates that the disparity in performance is most likely due to poor translation quality. We release our dataset and scripts at: \textcolor{Blue}{\url{https://github.com/Swati17293/KG-Multi-Bias}} for future research. Our framework has the potential to benefit journalists, social scientists, news producers, and consumers.

    \end{abstract}
    

\end{frontmatter}
\setlength{\parskip}{1em}

\section{Introduction}
\label{sec:introduction}

    News plays a significant role in the functioning of a democratic society \citep{helberger2019democratic,mcnair2009journalism}. Even though it is presumed to be a reliable source of information \citep{miller2000news}, bias is inevitable \citep{park2009newscube}. As a result, research communities devote a great deal of attention to the study of news bias \citep{davis2022gender,spinde2021interdisciplinary,spinde2021automated}. However, the first step in conducting such a study is to identify it \citep{chen2022partisan,chipidza2021effect}. Although the task may appear trivial, it is in fact challenging as bias can manifest itself at different levels in complex ways \citep{hamborg2019automated}. When it comes to news headlines, this task becomes even more challenging as headlines are inherently short, catchy or appealing, context-deficient, and contain only subtle bias clues \citep{gangula2019detecting,laban2021news}.

    With the rise of digital journalism and micro-blogging, the headline is becoming the only part of a news item that people read \citep{holmqvist2003reading}. Furthermore, since it serves as an entry point of an article, people are more likely to form an opinion by simply reading it without reading the rest of the article \citep{andrew2007media,ecker2014effects}. They seem to be swayed more by its creativity than its clarity \citep{molek2013towards}. Journalists often use this to their advantage by fabricating facts in a way that expresses their intended point of view, which captures the readers' emotions and interests \citep{andrew2007media,ifantidou2009newspaper}.
    
    Such biased reporting has a direct impact on how the public perceives events such as elections \citep{mccluskey2005content}, protests \citep{jovanovic2021headlines}, terrorism \citep{zeng2018critical}, and so on \citep{andrew2013political,navia2015mercurio}. Therefore, it is important to identify bias to help people form an unbiased and well-informed opinion \citep{hamborg2017identification,hamborg2020bias}. Some studies deal with news bias, but most of them are for High-Resource Languages (HRLs) such as English and German \citep{aksenov2021fine,guo2022modeling}. Such research is especially scarce for Low-Resource Languages (LRLs) \citep{doan2022survey}, even though mitigating the effects of bias is equally important in assisting readers of these languages \citep{park2012computational}.
    
    With a scarcity of standard labelled data, existing studies, and external knowledge to draw from, the task of news bias identification in these LRLs becomes even more challenging \citep{del2014lremap,doan2022survey}. As a result, resolving these issues necessitates understanding the narrative being presented \citep{bruneau2012going}. This can be accomplished by identifying connections between what is explicitly stated and what is implied \citep{berner1983commentary}.
    
    It is well-known that incorporating commonsense reasoning abilities can facilitate the inference of such connections by identifying a set of unstated causes and effects \citep{li2021enhancing,li2021past}. Such additional knowledge has been proven to be beneficial for several tasks \citep{du2022enhancing,li2021enhancing2,lieto2021commonsense}, including the prediction of bias in English news headlines \citep{swati4114271ic}. To this end, we use the popular neural knowledge model COMET \citep{hwang2021comet} trained on ATOMIC\rlap{\textsuperscript{20}}\textsubscript{20} \citep{hwang2021comet} to generate the Inferential Commonsense knowledge (IC\_Knwl). Since the textual descriptions of commonsense in the ATOMIC\rlap{\textsuperscript{20}}\textsubscript{20} knowledge repository are composed in English, it creates a language barrier.
    
\begin{figure*}[!htb]
        \centering
        \begin{subfigure}{.48\textwidth}
            \includegraphics[width=1.0\textwidth,height=0.65\textwidth]{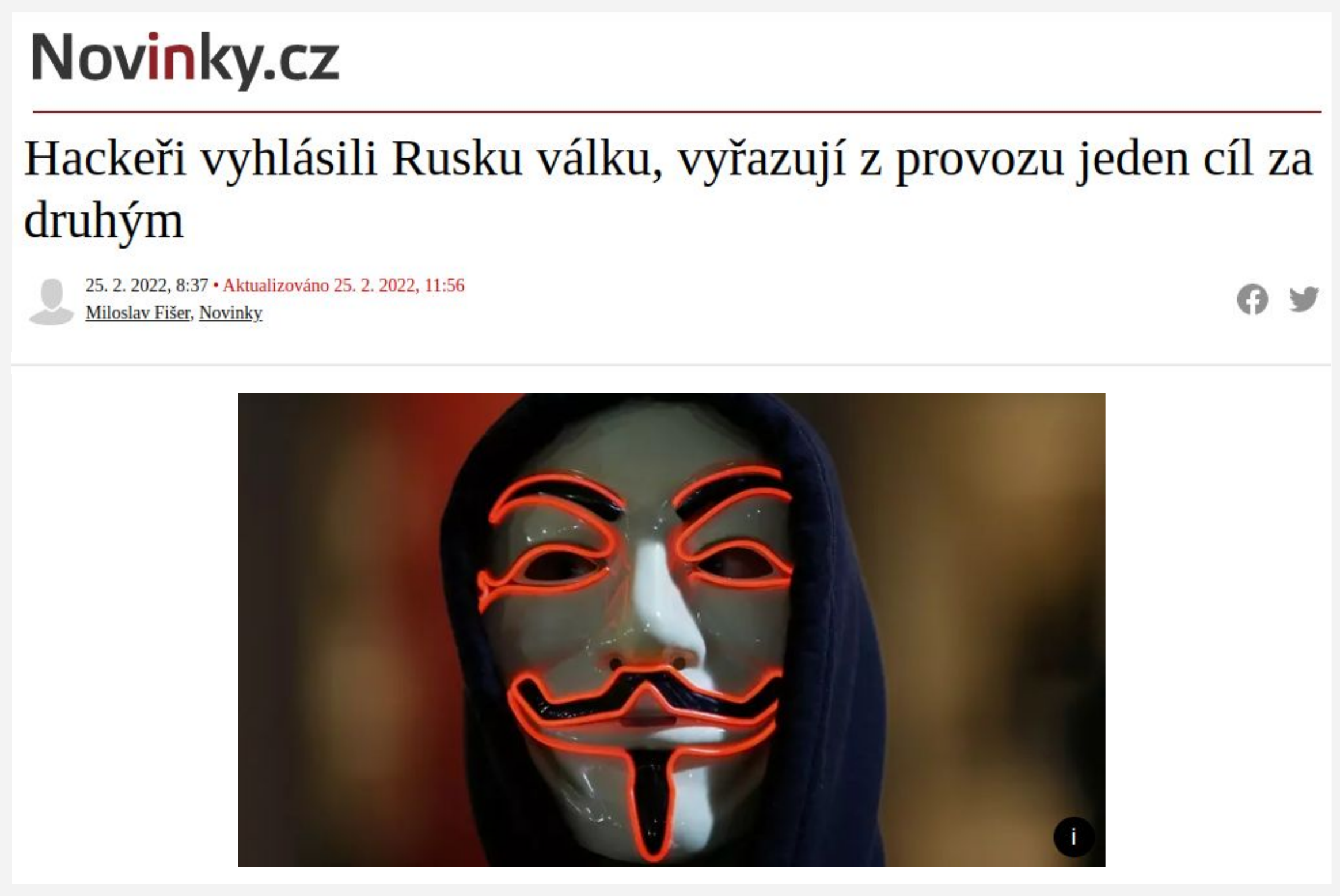}
            \caption{\textbf{Novinky.cz (Czech)}: Hackeři vyhlásili Rusku válku, vyřazují z provozu jeden cíl za druhým (\textcolor{Blue}{Hackers have declared war on Russia, decommissioning one target after another}) \\ \textbf{IC\_Knwl}: Hackeři jsou vidět jako `agresivní', který `chce zničit nepřítele` (\textcolor{Blue}{Hackers are \textit{seen as} `aggressive' who \textit{wants to} `to take revenge on Russia'})\\
            \textbf{Political polarity: \textcolor{Maroon}{Left Center}}}
            \label{fig:Czech-headline-lc}
        \end{subfigure}
        \hspace{0.5em}%
        \begin{subfigure}{.48\textwidth}
            \includegraphics[width=1.0\textwidth,height=0.65\textwidth]{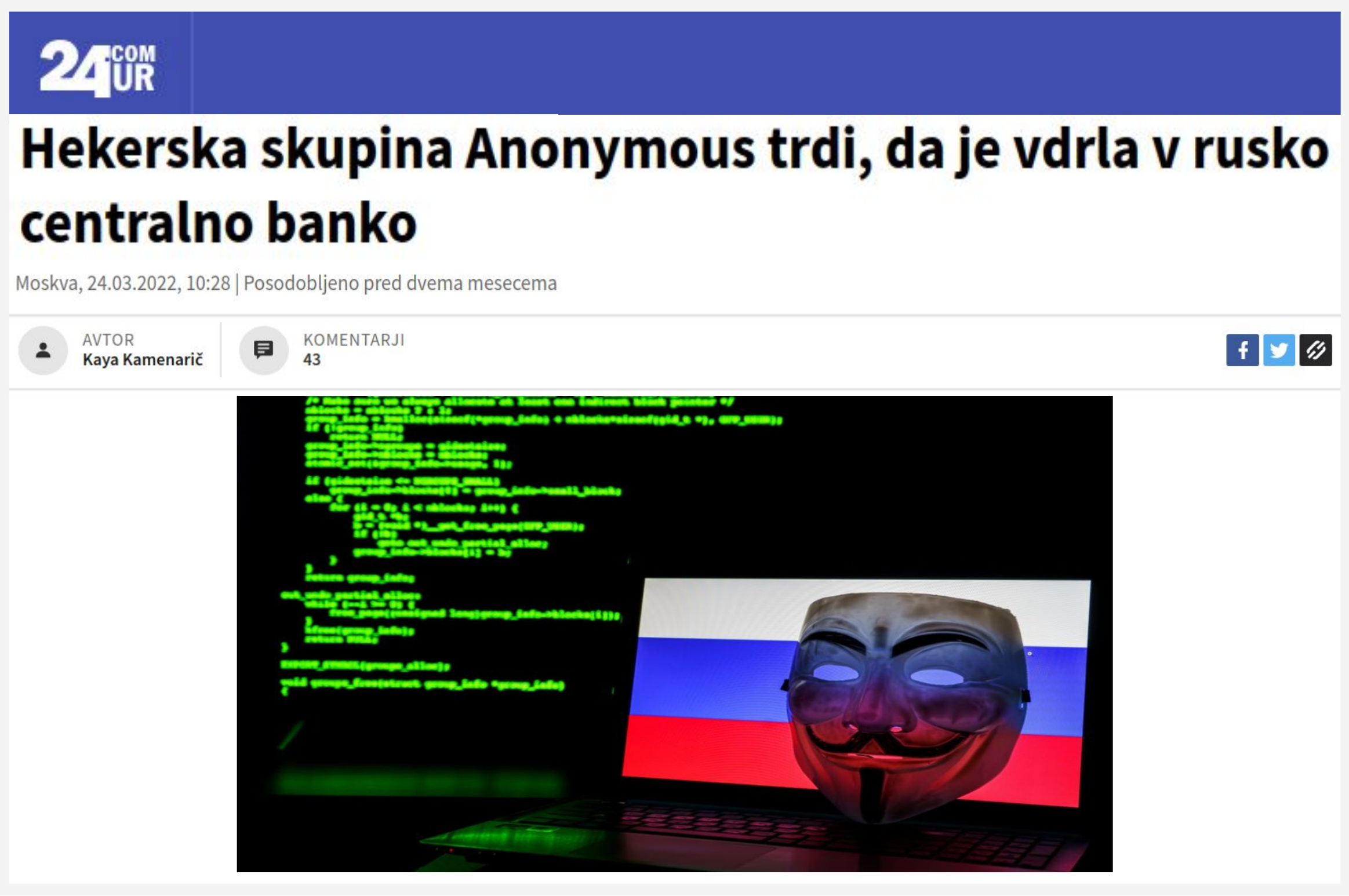}
            
            \caption{\textbf{24ur.com (Slovenian)}: Hekerska skupina Anonymous trdi, da je vdrla v rusko centralno banko (\textcolor{Blue}{The hacker group Anonymous claims to have hacked into Russia’s central bank}) \\ \textbf{IC\_Knwl}: Hekerji veljajo za `zlonamerne', ki želijo `dati izjavo' (\textcolor{Blue}{Hackers are \textit{seen as} `malicious' who \textit{wants to} `make a statement'})\\
            \textbf{Political polarity: \textcolor{Maroon}{Least Biased}}}
            \label{fig:Slovenian-headline-lb}
        \end{subfigure}
        \caption{News headlines from (a) Czech and (b) Slovenian news outlets on the ``\textit{hacker attacks on Russia}" with varying political polarities. Inferential Commonsense Knowledge (IC\_Knwl) can help improve prediction accuracy by facilitating the acquisition of additional bias-cues. \\ \scriptsize(Note: this example shows only a subset of IC\_Knwl relations. Image source: \href{https://www.24ur.com/novice/tujina/ukrajina/hekerska-skupina-anonymous-trdi-da-je-vdrla-v-rusko-centralno-banko.html}{24ur.com}, \href{https://www.novinky.cz/internet-a-pc/bezpecnost/clanek/hackeri-vyhlasili-rusku-valku-vyradili-z-provozu-statni-televizi-rt-40388303}{novinky.cz}, Translation: \href{https://translate.google.com/}{translate.google.com})}
        \label{fig:introduction-headline-bias}
    \end{figure*}

    Thus, to extend its capability beyond this barrier, we propose to leverage the Translate-Retrieve-Translate (TRT) approach \citep{fang2022leveraging}. Specifically, given a headline in the target language, TRT first translates it into English and then acquires the associated knowledge in English. It then translates the knowledge back into the target language. As illustrated in Figure \ref{fig:introduction-headline-bias}, IC\_Knwl in the target language can help enhance the prediction accuracy.
    
    To finally predict the political polarity of multilingual news headlines, we present a learning framework in Section \ref{sec:methodology}. Given a multilingual headline, we first utilise COMET with TRT to acquire IC\_Knwl in the target language. Next, we employ an attention mechanism to emphasise important inferences. We finally integrate the attended IC\_Knwl into a multilingual pre-trained language model for bias prediction.
    
    However, there are no standard labelled datasets available for evaluating our framework \citep{doan2022survey}. Prior studies either restrict their scope to news in a single language \citep{gangula2019detecting} or analyse news in different languages separately \citep{bonyadi2013headlines}. Even the overall ratings for news outlets that publish in these languages are unavailable on popular bias rating platforms such as \href{https://allsides.com/}{allsides.com} and \href{https://adfontesmedia.com/}{adfontesmedia.com}.
    
    Given the limited number of news outlets publishing in these LRLs for each bias class \citep{MBFCFinnish}, imbalanced data distribution poses another challenge. Furthermore, no labelled data may exist for some LRLs. Especially for European LRLs, data and knowledge resources are extremely scarce \citep{del2014lremap}. To this end, we present our dataset of news headlines in five European LRLs annotated with their respective political leanings (ref. Section \ref{sec:data-description}). It is constructed to mimic the challenges encountered by LRLs. 
    
    For a model to overcome the aforementioned challenges, cross-lingual transfer learning is crucial \citep{li2022cross,lu2015transfer,pamungkas2021joint}. It can be achieved with the help of multilingual Pre-trained Language Models (PLMs) \citep{feng2020language,reimers2019sentence,yang2021universal}. These models can generate vector embeddings of texts in different languages that are aligned in a single vector space, enabling few-shot/zero-shot learning. Advances in multilingual PLMs have shown promise in numerous NLP tasks \citep{kruspe2020cross,pei2022ab}. However, to use them effectively, systems must be fine-tuned to the task at hand \citep{patel2021efficient}. Unfortunately, as stated previously, the majority of these LRLs lack large enough data sets for such fine-tuning. They also suffer from the problem of specificity in their vocabulary that focuses on their cultural heritage, which further hinders the performance of these models \citep{talat2022you}. Therefore, in this study, we also evaluate several state-of-the-art multilingual PLMs for their effectiveness (ref. Section \ref{sec:feature-encoding}).
    
    \subsection{Contributions}
    \label{sec:contributions}
    
        The key contributions of our work are summarised as follows:
        \begin{itemize}
            \item {\textbf{}} Proposing to leverage Inferential Commonsense Knowledge (IC\_Knwl) through a Translate-Retrieve-Translate (TRT) strategy to facilitate comprehension of the overall narrative of the multilingual headlines.
            
            \item {\textbf{}}  Introducing an IC\_Knwl-infused language-agnostic learning framework for enhancing the prediction of political polarity in multilingual news headlines under imbalanced sample distribution. 
            
            \item {\textbf{}} Presenting a dataset of multilingual news headlines in five European low-resource languages annotated with their respective political polarities. 
            
            \item {\textbf{}} Thorough experiments with several state-of-the-art multilingual pre-trained language models to assess their effectiveness. 
            
            \item {\textbf{}} Analysing the impact of IC\_Knwl infusion on overall performance and across languages with and without attention mechanism.
            
        \end{itemize}
     
    The remainder of this paper is structured as follows: After a brief review of the key related works in Section \ref{sec:literature-review}, we introduce our dataset and provide an overview of its data collection framework in Section \ref{sec:data-description}. We then present the materials and methods utilised in this study in Section \ref{sec:materials-and-methods}. In Section \ref{sec:results-and-discussion}, we present the results and analysis of our experiments followed by research implications in Section \ref{sec:research-implications}. Finally, in Section \ref{sec:conclusions-and-future-works}, we present the concluding remarks and potential directions for future research.

\section{Literature review}
\label{sec:literature-review}

    In our learning framework, we predict the political polarity of multilingual news headlines by incorporating commonsense knowledge into a pre-trained multilingual language model. Consequently, we organise the related work in this section from these three perspectives as follows:
    
    \subsection{Prediction of polarity in multilingual news headlines}
    \label{sec:polarity-prediction-of-headlines}
    
        Researchers have long been interested in studying news articles and headlines in order to address problems such as fake news detection \citep{roy2019deep,saikh2019novel,saikh2019deep}, sentiment analysis \citep{king2021diffusion,rotim2017takelab}, topic modelling \citep{korenvcic2018document,pandur2020topic}, and so on \citep{muller2021multimodal,tahmasebzadeh2020feature}. While predicting the polarity of news articles is not a new problem \citep{d2012media,palic2019takelab,stevenson1980reconsideration}, modelling it at the headline level has received less attention \citep{swati4114271ic}. Earlier studies relied on predefined linguistic feature sets \citep{chen2018learning,groseclose2005measure} and standard machine learning techniques \citep{iyyer2014political}. Recent studies, on the other hand, have advanced to deep-learning techniques \citep{gangula2019detecting,naredla2022detection,tourni2021detecting}. In particular, Transformers-based models have demonstrated remarkable performance enhancements \citep{krieger2022domain,magotra2022news}. However, the majority of these studies focus on languages with abundant resources, with only a few exceptions studying languages with limited resources \citep{gangula2019detecting}. Moreover, these studies are either limited to a single language \citep{hoyer2016spanish,navia2015mercurio} or analyse news in different languages independently \citep{bonyadi2013headlines}.
        
        The lack of large-scale annotated gold-standard datasets for these languages further complicates the task \citep{doan2022survey,talat2022you}. Most existing datasets were generated manually \citep{gangula2019detecting}. Manual annotation requires a substantial amount of time and effort. Moreover, these small-scale datasets are not suitable for training deep learning models \citep{fan2019plain}. There are also datasets generated using an approach in the form of distant supervision, in which the polarity of a news outlet is mapped to each of its articles \citep{baly2020we,chen2018learning}. The polarity is typically obtained from prominent bias rating platforms, such as \href{https://allsides.com/}{allsides.com} and \href{https://adfontesmedia.com/}{adfontesmedia.com} where a team of domain experts employs specialised guidelines for annotations. Even though distant supervision facilitates the creation of large datasets, bias ratings are typically not available for all outlets, especially those that publish in languages with limited resources \citep{MBFCFinnish}. Another possibility is to combine the datasets available in different languages. However, this strategy would result in an uneven distribution of topics and events across polarity classes and languages.
        
        To mitigate the aforementioned issues of data scarcity, we present a diverse and scalable multilingual news headline dataset in five low-resource languages to predict political leanings (ref. Section \ref{sec:data-description}). Inspired by but distinct from these related works, we then introduce our learning framework (ref. Section \ref{sec:methodology}). We infuse it with inferential commonsense knowledge and explore its application for the task of polarity prediction. Furthermore we propose a language-agnostic learning framework which we utilise to evaluate the effectiveness of several state-of-the-art multilingual pre-trained language models.
        
        \subsection{Commonsense knowledge}
        \label{sec:commonsense-knowledge}
    
            Multiple studies have revealed that large-scale pre-trained language models are implicitly capable of encoding some commonsense and factual knowledge \citep{petroni2019language,shwartz2020unsupervised}. However, these models hardly acquire inferential commonsense knowledge, especially in context-deficient settings \citep{do2021rotten,kassner2020negated}. Consequently, recent studies have investigated the application of such knowledge in a number of  NLP-related tasks \citep{tu2022misc,young2018augmenting,zhou2018commonsense}. It has been demonstrated that injecting such knowledge improves output performance on a variety of tasks, including reading comprehension \citep{mihaylov2018knowledgeable}, question answering \citep{lal2022analyzing}, and story generation \citep{chen2019incorporating}, among others \citep{wang2013common,zhong2021care,zhu2021topic}.
        
            There exist several widely used commonsense knowledge resources such as ConceptNet \citep{speer2017conceptnet}, SentiNet \citep{cambria2018senticnet}, GLUCOSE \citep{mostafazadeh2020glucose}, ATOMIC\rlap{\textsuperscript{20}}\textsubscript{20} \citep{hwang2021comet}, etc \citep{rashkin2018event2mind,romero2020inside,tandon2017webchild}. ConceptNet is a semantic network containing concept-level relational commonsense knowledge as phrases and words in natural language. SentiNet is a well-known resource used for sentiment analysis at the concept level. GLUCOSE is a large-scale resource used for capturing implicit casual knowledge in narrative contexts. Structured as if-then relations with an emphasis on inferential knowledge, ATOMIC\rlap{\textsuperscript{20}}\textsubscript{20} is a resource composed of everyday commonsense knowledge. 
            
            These knowledge resources are used to train generative models such as COMET \citep{hwang2021comet} and ParaCOMET \citep{gabriel2021paragraph}. Trained on ConceptNet and ATOMIC\rlap{\textsuperscript{20}}\textsubscript{20}, COMET is capable of generating a diverse range of context-relevant commonsense descriptions. Motivated by the related studies, we thus use COMET trained on the ATOMIC\rlap{\textsuperscript{20}}\textsubscript{20} knowledge base. However, different from these studies, we use it to identify unstated causes and effects in context-deficient headlines.
            
        \subsection{Multilingual pre-trained language models}
        \label{sec:multilingual-language-models}
    
            A number of language representation models, such as BERT \citep{kentonbert}, ELECTRA \citep{clark2020electra}, XLNet \citep{yang2019xlnet}, etc. \citep{lan2019albert,peters2018deep}, have emerged in recent years. The majority of them are based on transformers, a non-sequential deep learning approach that provides positional embeddings via a multi-headed attention technique \citep{vaswani2017attention}. Due to their many advantages \citep{lin2021survey}, they are popular not only for solving a wide range of NLP-related tasks \citep{gain2021iitp,mishra2022please,yadav2021nlm,yadav2022question} but also for a variety of other practical applications \citep{pingali2021multimodal,shin2019effective,singh2022unity}.
        
            A number of their multilingual variants, such as Multilingual BERT (mBERT) \citep{kentonbert}, XLM-RoBERTa (XLM-R) \citep{conneau2020unsupervised}, and Multilingual Bidirectional Auto- Regressive Transformers (mBART) \citep{liu2020multilingual}, have shown promising results for text processing across multiple languages \citep{li2022cross,kumar2021sentiment,novak2022document}. If followed by task-specific fine-tuning, they have proven to be effective \citep{muller2021first}. However, they are ineffective at generating sentence-level representations \citep{cer2018universal}.
            
            Several models designed to generate semantically meaningful sentence representations, such as Sentence BERT (SBERT) \citep{reimers2020making}, Universal Sentence Encoder (USE) \citep{cer2018universal}, and Language-Agnostic Sentence Representations (LASER) \citep{feng2022language}, were proposed to address this limitation. They have proven useful in a variety of NLP applications \citep{chaudhary2019low,mohammad2021gated}. Over the past few years, several similar frameworks have been extended to support over 100 languages \citep{cer2018universal,conneau2017supervised}. Some even support low-resource languages such as Slovenian, Romanian, and so on \citep{feng2022language,yang2021universal}.
            
            Despite having millions of parameters and being trained on diverse datasets, these models are not guaranteed to generalise to all tasks and domains \citep{muller2021first}. As a result, we investigate and compare several state-of-the-art PLMs in this study for their effectiveness.

\section{Dataset}
\label{sec:data-description}
    We introduce our dataset and describe its data collection framework in this section. To begin with, we introduce two primary data sources that serve as the foundation for our dataset. We then present a detailed description of our framework for data collection followed by a description of our dataset.
    
    \subsection{Primary data sources}
    \label{sec:primary-data-sources}
        We present two primary data sources Media Bias/Fact Check (MBFC) and Event Registry (ER) in this section. We use the bias rating portal MBFC to select media outlets and retrieve their associated bias labels. We use ER to crawl the headlines of articles published by these selected media outlets.
        
        \subsubsection{Media Bias Fact/Check}
        \label{sec:media-bias-fact-check}
            Several well-known platforms, such as \href{https://allsides.com/}{allsides.com}, \href{https://adfontesmedia.com/}{adfontesmedia.com}, and \href{https://mediabiasfactcheck.com}{mediabiasfactcheck.com} \citep{MBFC}, publish bias ratings for media outlets. However, due to the scarcity of such ratings for outlets in low-resource languages, we choose to acquire labels exclusively from mediabiasfactcheck (MBFC). It is a trustworthy bias rating and fact-checking platform with extensive coverage and regular updates. It has been employed to predict and assess media bias in a number of studies \citep{baly2018predicting}. In addition, it has also been utilised to develop tools such as `Iffy Quotient' \citep{resnick2018iffy}, which monitors the prevalence of fake news and questionable sources on social media.
            
            To assign bias ratings to media sources, it establishes five levels of political bias: `left', `left-center', `center', `right-center', and `right' \citep{MBFCMethodologyBIAS}. It also assigns ratings based on their credibility and factual accuracy. These ratings are assigned by a group of paid contractors and volunteers who are instructed to adhere to a predetermined methodology \citep{MBFCMethodology}. Based on a quantifiable system, its methodology includes both objective and subjective measures.
            
        \subsubsection{Event Registry}
        \label{sec:event-registry}
            To scrape news headlines, we use the Event Registry \citep{LebanGregorEventRegistry} platform. It has a custom collection of over 150,000 diverse sources from around the world in over 50 languages. It is widely used in studies involving news event analysis \citep{swati2021eveout,mladenic2020you,mladenic2021understanding}. Its primary objective is to cluster contents as events, but it also facilitates the collection of news stories and articles. It offers a Python API\footnote{\url{https://eventregistry.org}} for accessing news content minutes after it has been published online. It has several search options for filtering out the desired content, such as searching by any news outlet, keyword, language, and among others. Using this API it is possible to extract news content as well as metadata published by different publishers in different languages. 
        
    \subsection{Data collection framework}
    \label{sec:data-collection-framework}
        
        \begin{figure*}[htb]
            \centering
            \includegraphics[width=1.0\textwidth,height=0.65\textwidth]{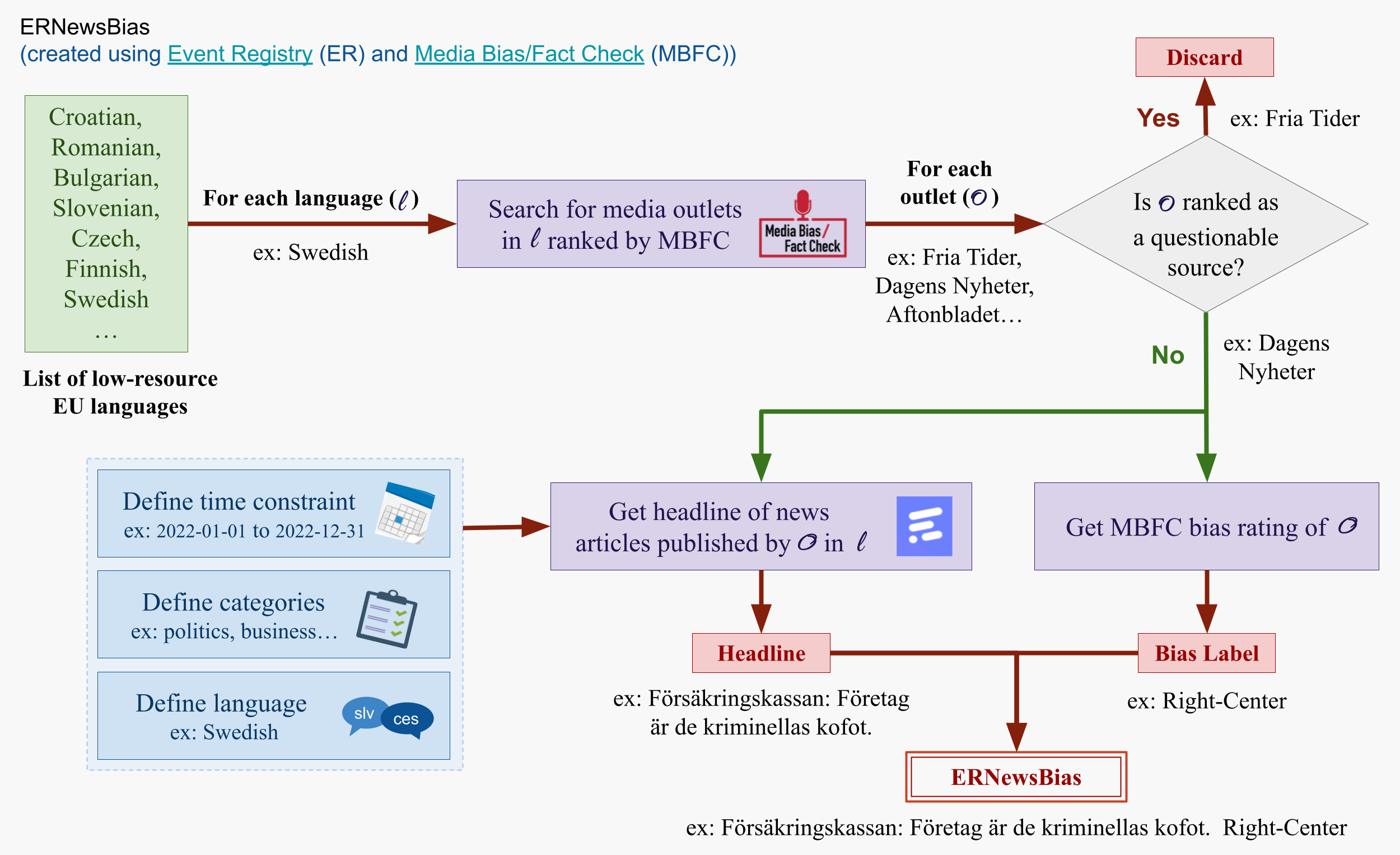}
            
            \caption{Data Collection Framework. We use Media Bias/Fact Check (MBFC) and Event Registry (ER) as the primary data sources in the framework.}
            \label{fig:data-collection-framework}
        
        \end{figure*}
        
        As illustrated in the data collection framework in Figure \ref{fig:data-collection-framework}, we begin the process by compiling a list of low-resource European languages (L). $L=\{l_1,l_2,...l_n\}$, with $n$ representing the total number of languages in the list. $\forall l \in L$, we then compile a list of media outlets ($O$) publishing in $l$ ranked by MBFC (ref. Section \ref{sec:media-bias-fact-check}). We define $O=\{o_1,o_2,...o_m\}$, with $m$ as the the total number of outlets in the list. $\forall o \in O$, we then check whether $o$ is ranked as a questionable source or not. Since questionable sources are prone to promote unfounded claims or theories as facts and offer little or no references to credible sources of information, they may turn out to be untrustworthy. Therefore, we discard such sources. $\forall$ unquestionable $o$, we extract the political bias label $b$ assigned by MBFC.
        
        We then define an explicit temporal query ($Q_{t}$):
        \begin{eqnarray}
        \label{eq:temporal-query}
            & Q_{t} = \{Q_{o}, \; Q_{l}, \; Q_{cat}, \; Q_{dt}\}
        \end{eqnarray}
        Where, $Q_{o}$, $Q_{l}$, and $Q_{cat}$ defines the query $o$, $l$, categories\footnote{\url{https://eventregistry.org/documentation?tab=suggCategories} Note: For our dataset, we only use the categories defined by ER as `news'.} respectively, and $Q_{dt}$ defines the time-constraint using $Q_{sd}$ and $Q_{ed}$ as the start and end dates:
        \begin{eqnarray}
        \label{eq:time-constraint}
            Q_{dt} = [Q_{sd},Q_{ed}]
        \end{eqnarray} 
        To scrape all the article headlines ($H$) published by each unquestionable $o$, we utilise $Q_{t}$ to query the Event Registry (ER) (ref. Section \ref{sec:event-registry}):
        \begin{eqnarray}
        \label{eq:news-headline-crawl}
            H = ER\;(Q_{t})
        \end{eqnarray}
        Finally, we assign the previously extracted bias label $b$ to the headlines in $H$ to construct the dataset. To generate the train/valid/test splits, we adopt a stratified split to simulate the imbalance in the collected data across the languages.
        
   \subsection{Dataset description}
   \label{sec:dataset-description} 
        Our dataset consists of news headlines annotated with their respective political leanings. We construct it to mimic the challenges encountered by LRLs. We begin by selecting five low-resource European languages: \textit{Czech}, \textit{Finnish}, \textit{Romanian}, \textit{Slovenian}, and \textit{Swedish}. We then compile a list of media outlets ranked by MBFC in these selected languages. We end up with seven news outlets: \textit{24ur}, \textit{Dagens Nyheter}, \textit{Delo}, \textit{Digi24}, \textit{Helsingin Sanomat}, \textit{Hotnews}, and \textit{Novinky} with bias labels: \textit{Left Center}, \textit{Least Biased}, and \textit{Right center}. In the end, we manage to generate $62,689$ news headlines with an average length of $10.2$ words. 

        In Table \ref{table:dataset-statistics}, we list the statistics for each language in the dataset. It is carefully documented and adheres to the requirements of the FAIR Data Principles\footnote{\url{https://www.nature.com/articles/sdata201618/}}. 
            
            \begin{table*}[htb]
                \centering
                \small
                  
                    \begin{tabular}{l l l l l l l}
                    
                    {} & \textbf{All} & \textbf{Czech} & \textbf{Finnish} & \textbf{Romanian} & \textbf{Slovenian} & \textbf{Swedish} \\ \hline

                    \textbf{Train} & 50,157 & 9,992 & 7,120 & 5,829 & 15,557 & 11,659 \\ 
                    
                    \textbf{Test} & 6,269 & 1,237 & 940 & 756 & 1,879 & 1,457 \\ 
                    
                    \textbf{Valid} & 6,263 & 1,310 & 880 & 764 & 1,853 & 1,456 \\ 
                    
                    \textbf{Total} & 62,689 & 12,539 & 8,940 & 7,349 & 19,289 & 14,572 \\ 
                    
                    \textbf{Len.} & 10.2 & 9.4 & 10.2 & 12.8 & 8.8 & 8.9 \\ \hline
                    
                    \end{tabular}
                
                \caption{Dataset Statistics. Len: average number of words in the headline.}
                \label{table:dataset-statistics}
                
            \end{table*}
          
\section{Materials and methods}
\label{sec:materials-and-methods}

    In this section, we begin by stating the research objectives followed by formally defining the task of predicting the political polarity of multilingual news headlines. We then present our learning framework and its key components, followed by a brief discussion of baseline models and the evaluation metrics used in this study. 
    
    \subsection{Research objectives}
    \label{sec:research-objectives}
    
        The primary objective of this study is to investigate the impact of our proposed framework for predicting political polarity in multilingual news headlines. It takes the advantage of the state-of-the-art pre-trained language models and the inferential commonsense knowledge in a multilingual setting. In this context, we define the following research objectives :
        
        \begin{itemize}
            
            \item{\textbf{RO1: }Introduce a knowledge-infused language-agnostic learning framework.}
            
            \item{\textbf{RO2: }Evaluate the impact of using an inferential commonsense knowledge as a source of additional information in a multilingual setting.}
            
            \item{\textbf{RO3: }Compare the effectiveness of several state-of-the-art multilingual pre-trained language models.}
            
            \item{\textbf{RO4: }Investigate the influence of knowledge attention on prediction performance.}
            
        \end{itemize}
        
    \subsection{Task definition}
    \label{sec:task-definition}
    
        We denote a language by $l \in L$, a short news headline text by $H$, an auxiliary piece of information as inferential commonsense knowledge by $IC\_Knwl$, a $H$ in $l$ as $H^l$, an $IC\_Knwl$ in $l$ as $IC\_Knwl^l$, and a political bias label by $b \in B$. We define the sets $L=\{l_1, l_2, ... l_n\}$ and $B=\{b_1, b_2, ... b_N\}$, where $n$ and $N$ represent the number of languages and bias labels in the respective sets $L$ and $B$. Given $H^l$, its corresponding $IC\_Knwl^l$ can be acquired using the commonsense knowledge modelling function $C$ with the appropriate model parameters $\alpha$, as shown in Eq. \ref{eq:IC_Knwl_acquisition}.
        \begin{eqnarray}
        \label{eq:IC_Knwl_acquisition}
            IC\_Knwl^l = C(H^l, \alpha)
        \end{eqnarray}
        $H^l$ can then be fused with the acquired $IC\_Knwl^l$ to represent its extended feature space $(H^l,IC\_Knwl^l)$. Given $H^l$, the task aims to train a classifier that maps its extended feature space to the bias set $B$. It can be mathematically formulated using Eq. \ref{eq:task_definition} with $f$ as the bias prediction function and $\theta$ as the model parameters.
        \begin{eqnarray}
        \label{eq:task_definition}
          b = f((H^l, IC\_Knwl^l), \theta)
        \end{eqnarray}
        
    \subsubsection{Methodology}
    \label{sec:methodology}
        
\begin{figure*}[htb]
            \centering
            \includegraphics[width=1.0\textwidth,height=0.95\textwidth]{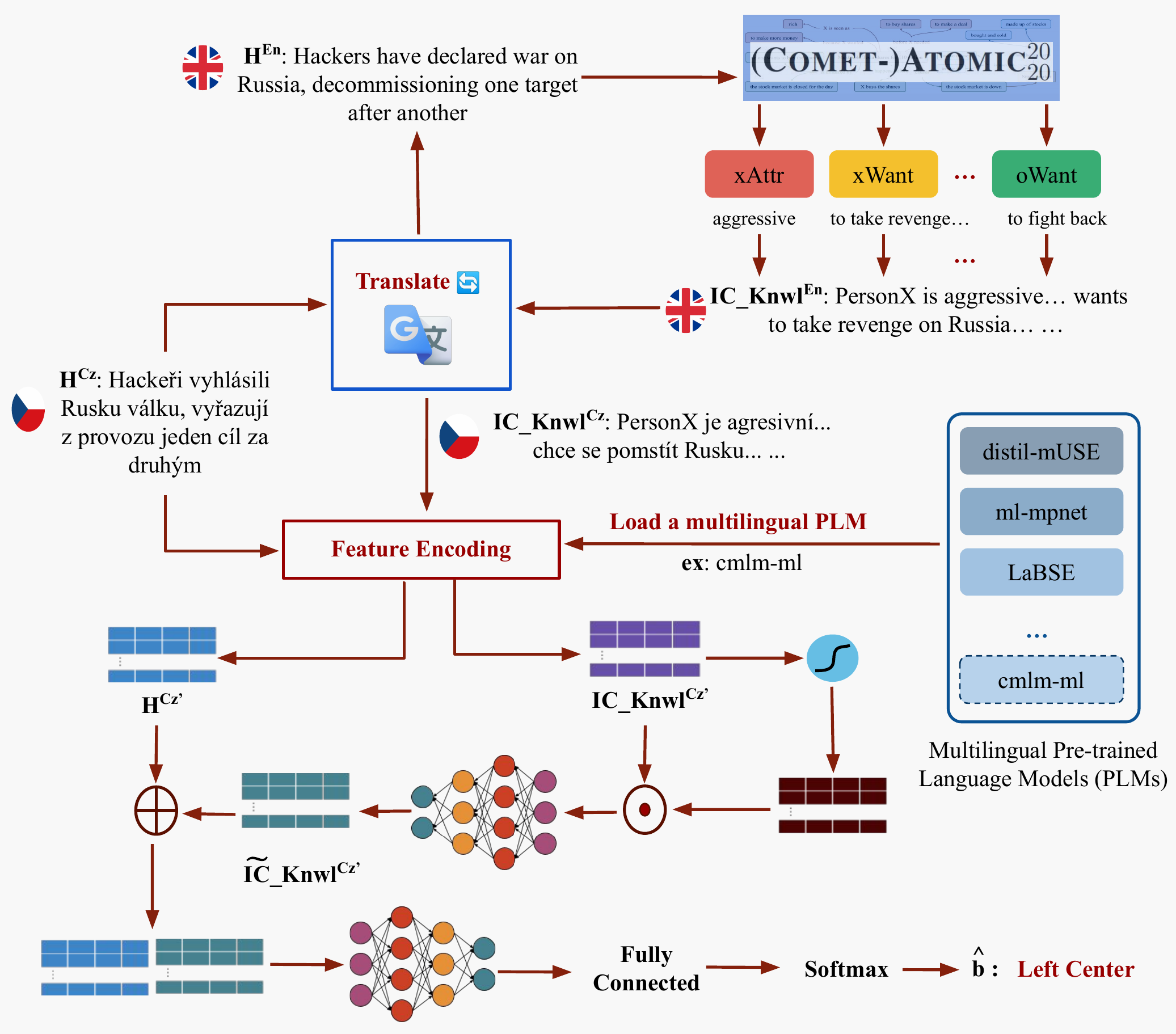}
            
            \caption{ An overview of our proposed learning framework. To predict political polarity of multilingual news headlines, it combines Inferential Commonsense Knowledge retrieved via the Translate-Retrieve-Translate strategy with multilingual pre-trained language models.}
            \label{fig:framework-overview}
        
        \end{figure*}

        To fulfill \textbf{RO1}, we propose a framework which is primarily based on inferential commonsense knowledge. It helps uncover contextual features that in turn can help predict the polarity of multilingual news headlines. To facilitate generalisation, our framework is compatible with any multilingual pre-trained language model. Figure \ref{fig:framework-overview} depicts its overall architecture. Its key components include Knowledge Acquisition, Feature Encoding, Knowledge Attention, and Bias Prediction. Each of these components is described in detail in the following subsections.

        \subsubsection{Knowledge acquisition}
        \label{sec:knowledge-acquisition}
        
\begin{figure*}[htb]
                \centering
                \includegraphics[width=1.0\textwidth,height=0.31\textwidth]{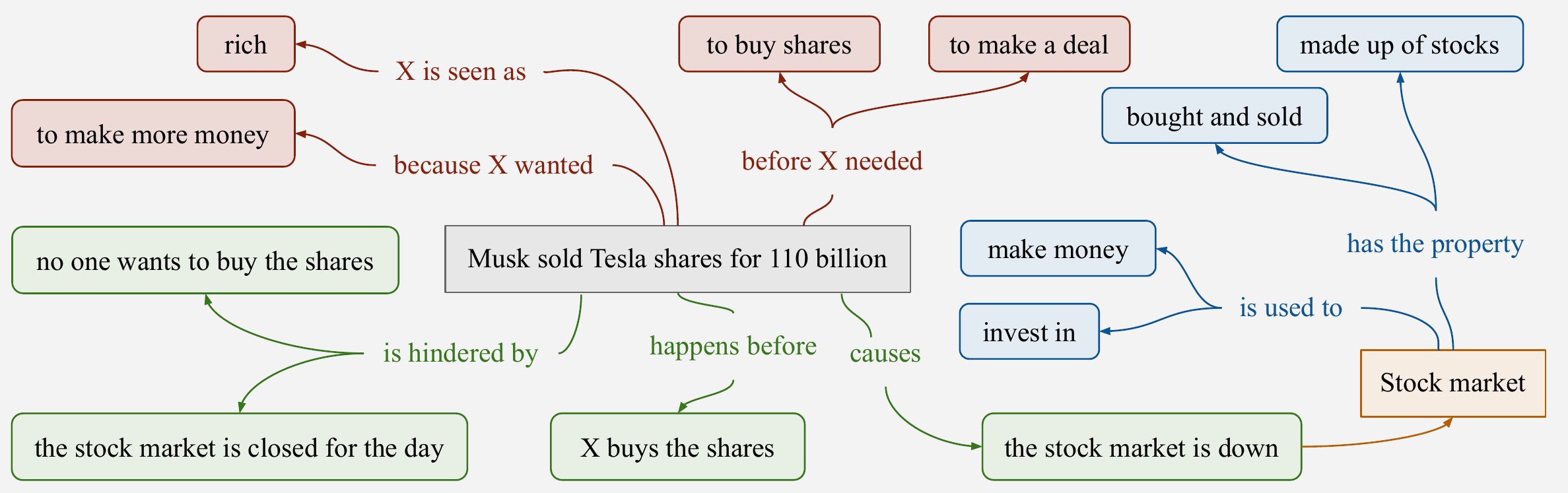}
                
                \caption{A small subset of $IC\_Knwl$ relations generated using $\text{ATOMIC}^{20}_{20}$ as the knowledge base in response to the news headline `\textit{Musk sold Tesla shares for 110 billion}'. Nodes in the colours red, green, blue, and orange represent relations depicting social interactions, events, physical entities, and category intersection, respectively. }
                \label{fig:IC_Knwl}
                
            \end{figure*}

            The $\text{ATOMIC}^{20}_{20}$ (ATlas Of MachIne Commonsense 2020)\footnote{\url{https://allenai.org/data/atomic-2020}} \citep{hwang2021comet} is a well-known, publicly available commonsense knowledge resource that is ``\textit{able to cover more correct facts about more diverse types of commonsense knowledge than any existing, publicly-available commonsense knowledge resource}". Its relations are composed of textual descriptions containing more than one million tuples of everyday inferential knowledge about entities and events. It is coded into different relation types, which are categorised into different sub-types, such as nine commonsense relations for social interaction, seven for physical entities, and seven for events. Figure \ref{fig:IC_Knwl} illustrates a subset of these relations generated in response to a sample news headline. 
            
            Relations of type social-interaction provide an insight into socially triggered states and behavioural patterns. As demonstrated by the examples in Table \ref{table:IC_Knwl-relations}, it is valuable for predicting people's reactions and behaviour in a given situation by assessing their intentions and goals. Motivated by its effectiveness in enhancing the performance of models designed to handle short news headlines in English language \citep{swati4114271ic}, we utilise it as the sole relation type for $IC\_Knwl$ in our work.
            
            \begin{table*}[htb]
                \centering
                \small
                
                \begin{tabular}{l l l}
                
                \textbf{Relation} & \textbf{Interpretation} & \textbf{Examples}\\ \hline
                
                {xAttr} & {X is seen as} & {lucky; competitive} \\
                
                {xEffect} & {as a result, X } & {wins the game; personx wins the race} \\
                
                {xIntent} & {because X wanted} & {to win; to be the best} \\
                
                {xNeed} & {but before, X needed} & {to train hard; to enter the contest} \\
                
                {xReact} & {as a result, X feels} & {happy; excited} \\
                
                {xWant} & {as a result, X wants} & {to celebrate; to win} \\
                
                {oEffect} & {as a result, others} & {loses the game; loses money} \\
                
                {oReact} & {as a result, others feel} & {disappointed; sad} \\
                
                {oWant} & {as a result, others want} & {to congratulate X, to win the game} \\ \hline
                
                \end{tabular}
                
                \caption{Examples of social interaction relation retrieved using $\text{ATOMIC}^{20}_{20}$ as the knowledge base for the short news headline `Grit Won'. Each relation type is interpreted using the human-readable template provided in \citep{hwang2021comet}.}
                \label{table:IC_Knwl-relations}
            
            \end{table*}

            To retrieve $IC\_Knwl$, we use COMmonsensE Transformers (COMET) \footnote{\url{https://github.com/allenai/comet-atomic-2020/}} \citep{bosselut2019comet,hwang2021comet} trained on the $\text{ATOMIC}^{20}_{20}$ knowledge graphs. COMET is a large pre-trained neural-network model that generates $IC\_Knwl$ in response to a query text. Given $H$, Inference type ($I_{type}$), and number of returned references ($k$), $IC\_Knwl$ can be retrieved using the following equation,
            \begin{eqnarray}
                \label{eq:comet-acquisition}
                    IC\_Knwl = COMET(H,I_{type},k)
                \end{eqnarray} 
            where, $I_{type}=[i_1, i_2, ... i_x]$ with $i$ as the inference type defined in Table \ref{table:IC_Knwl-relations} and $x$ as the total relations in the set. Since COMET returns the $IC\_Knwl$ as a list of inference results $\forall i \in I_{type}$, we set $k=1$ to return only one inference result per $I_{type}$. Furthermore, while retrieving $IC\_Knwl$, we combine the returned pieces of inferences of each $I_{type}$ to make it more meaningful. For example, 
            \begin{enumerate}[\indent {}]
                \item \textbf{Headline:} \textcolor{Blue}{\textit{Grit Won}}
                \item \textbf{IC\_Knwl:} xAttr: \textcolor{Blue}{\textit{lucky}}, xIntent: \textcolor{Blue}{\textit{to win}}, xEffect: \textcolor{Blue}{\textit{wins the game}}, xWant: \textcolor{Blue}{\textit{to celebrate}}, xReact: \textcolor{Blue}{\textit{happy}}, oWant: \textcolor{Blue}{\textit{to congratulate X}}, oEffect: \textcolor{Blue}{\textit{looses the game}}, oReact: \textcolor{Blue}{\textit{disappointed}}
                \item \textbf{Processed IC\_Knwl:} PersonX is \textcolor{Blue}{\textit{lucky}}, needed \textcolor{Blue}{\textit{to train hard}}, intended \textcolor{Blue}{\textit{to win}}, \textcolor{Blue}{\textit{wins the game}}, wants \textcolor{Blue}{\textit{to celebrate}}, feels \textcolor{Blue}{\textit{happy}}. Others want \textcolor{Blue}{\textit{to congratulate X}}, \textcolor{Blue}{\textit{looses the game}}, feel \textcolor{Blue}{\textit{disappointed}}.
            \end{enumerate}
            
            Finally, to generate $IC\_Knwl^l$ for $H^l$, we use the aforementioned method along with the Translate-Retrieve-Translate (TRT) approach \citep{fang2022leveraging}. Specifically, given a $H^l$, we first translate it into English and retrieve its associated $IC\_Knwl$ in English. We then translate the retrieved $IC\_Knwl$ into the target language $l$ to finally get the $IC\_Knwl^l$. We use the Google Translate API\footnote{\url{https://cloud.google.com/translate}} for our translations.

        \subsubsection{Feature encoding} 
        \label{sec:feature-encoding}
        
            To acquire feature vectors $H^l{}'$ and $IC\_Knwl^l{}'$, we use multilingual pre-trained language models (PLMs). For their optimal performance, they are required to map embedding vectors of text written in different languages into a single vector space. As a result, the degree of vector alignment influences their performance. In this regard, we explore the state-of-the-art multilingual PLMs defined in Section \ref{sec:baseline-models}. These PLMs differ from word-embedding models as they are trained on a wide range of tasks that require modelling the meaning of word sequences as opposed to individual words.
            
        \subsubsection{Knowledge attention}
        \label{sec:knowledge-attention}
        
            Ideally, not all retrieved inferences are expected to be of the same relevance. Consequently, we use the Sigmoid function \citep{nantomah2019some} on $IC\_Knwl^l{}'$ to determine the relevance of each of them. Following the work of Majumder et al. \citep{majumder2020mime}, we then multiply $IC\_Knwl^l{}'$ by the resulting relevance scores to highlight the most significant inferences. We use this vector in a Multi-Layer Perceptron (MLP) network trained to mix inferences from different $I_{type}$ to finally generate the attended vector $\widetilde{IC}\_Knwl^l{}'$:
                \begin{eqnarray}
                \label{eq:knowledge_attention}
                    \widetilde{IC}\_Knwl^l{}' = MLP(Sigmoid(IC\_Knwl^l{}') \;\odot\; IC\_Knwl^l{}')
                \end{eqnarray} 
                where $\odot$ denotes element-wise multiplication.

        \subsubsection{Bias prediction}
        \label{sec:polarity-prediction}
           
           To predict the bias label $\hat{b}$, we first fuse the vectors $H^l{}'$ and $\widetilde{IC}\_Knwl^l{}'$ to generate $F$: 
            \begin{eqnarray}
            \label{eq:feature-fusion}
                F = H^l{}' \;\oplus\; \widetilde{IC}\_Knwl^l{}'
            \end{eqnarray} 
            where $\oplus$ represents the concatenation operation. 
            
            We then feed the fused vector $F$ to an MLP network and forward the resultant vector to a Fully Connected layer (FC) having Softmax ($\sigma$) activation to finally predict $\hat{b}$:
            \begin{eqnarray}
            \label{eq:bias-prediction}
                \hat{b} = FC(\sigma(MLP(F))
            \end{eqnarray} 
            
            We train our network using the AdaMax \citep{kingma2014adam} as the optimizer with its default parameters. We use the Categorical cross-entropy as the loss function, which is defined as follows:
            \begin{eqnarray}
                \label{eq:categorical-cross-entropy}
                Loss = -\sum_{i = 1}^{|B|} (b_i*\log(\hat{b}_i))
            \end{eqnarray} 
            where $b_i$ and $\hat{b}_i$ are the actual and predicted probabilities of selecting the $i^{th}$ bias label in $B$.
            
    \subsection{Baseline models}
    \label{sec:baseline-models}
    
        Based on their superior performance in a variety of related tasks in multilingual settings \citep{pei2022ab,sayar2020leveraging}, we chose the following state-of-the-art baseline models for a comprehensive evaluation of our proposed framework.
        
        \begin{itemize}
        
            \item \textbf{ml-MiniLM} \citep{reimers2020making} [paraphrase-multilingual-MiniLM-L12-v2]: a multilingual version of the sentence transformer, paraphrase-MiniLM-L12-v2 \citep{reimers2019sentence}. It generates 384-dimensional aligned dense vectors. It is pre-trained on parallel data for more than 50 languages. It trades accuracy for speed and its reduced dimension results in lower memory requirements.

            \item \textbf{distil-mUSE} \citep{reimers2020making} [distiluse-base-multilingual-cased-v2]: multilingual Universal Sentence Encoder (mUSE) \citep{yang2020multilingual} is based on the transformer architecture \citep{vaswani2017attention}, which uses a multi-task trained dual-encoder to embed texts into a single vector space. The multilingual knowledge distilled version of mUSE (distil-mUSE) supports over 50 languages. It maps text to a 512-dimensional dense vector space. 
            
            \item \textbf{ml-mpnet} \citep{reimers2020making} [paraphrase-multilingual-mpnet-base-v2]: a multilingual version of the sentence transformer, paraphrase-mpnet-base-v2 \citep{reimers2019sentence}. It is pre-trained on parallel data for over 50 languages and generates 768-dimensional aligned dense vectors. It outperforms other multilingual models based on sentence transformers. However, its increased computational complexity makes it time-intensive.
            
            \item \textbf{LaBSE} \citep{feng2020language}[LaBSE/2]: a language-agnostic BERT based model that maps text into a 768-dimensional dense vector space. To map single plain-text segments to encoder inputs, it requires a separate preprocessor API build for the universal-sentence-encoder-cmlm multilingual models\footnote{\scriptsize{\url{https://tfhub.dev/google/universal-sentence-encoder-cmlm/multilingual-preprocess/2}}}. It is trained and optimised to generate aligned vectors for bilingual sentence pairs, and it currently supports over 109 languages. Although the model, like other BERT models, can be fine-tuned, the authors recommend that it be used as it is.
            
            \item \textbf{cmlm-ml} \citep{yang2021universal} [cmlm/multilingual-base/1]: a multilingual model trained with a conditional masked language model (cmlm-ml). Its architecture is based on a 12-layer BERT transformer \citep{devlin2018bert}, but it is far more complex. Similar to LaBSE, it also requires an additional preprocessor to map plain-text inputs to encoder inputs. It transforms text into 768-dimensional aligned vectors and supports more than 100 languages. Although its inference speed is significantly slower than that of other comparable models, its performance is far superior.
                
        \end{itemize}
        
    \subsection{Evaluation metrics}
    \label{sec:evaluation-metrics}
    
        To assess the performance of our proposed framework, we employ well-known metrics used to evaluate prediction models \citep{kowsari2019text}, such as Accuracy (A) and F\textsubscript{1}-score ($F_{1}$). However, in the case of an imbalanced dataset like ours, where true negative instances outnumber true positive instances for several languages, they are not a reliable indicator. Jaccard (J) \citep{real1996probabilistic} score is a reliable metric for evaluating models where no examples exist for each class. It disregards true negatives in favour of true positives, facilitating the interpretation of the results. It is even more reliable when evaluating models for individual languages since the imbalance is more apparent. As a result, we also employ the Jaccard score to gain a deeper understanding. We compute these metrics using the values of the confusion matrix defined in Table \ref{table:confusion-matrix}. 
        
        \begin{table*}[htb]
            \centering
            \small
              
            \begin{tabular}{l l}
            
            \hline
            \textbf{True Positive (TP):} & {label is present and is predicted.} \\
            
            \textbf{True Negative (TN):} & {label is not present and is not predicted.} \\
            
            \textbf{False Positive (FP):} & {label is not present but is predicted.} \\
            
            \textbf{False Negative (FN):} & {label is present but is not predicted.} \\ \hline

            \end{tabular}
            
            \caption{Description of the values of the confusion matrix.}
            \label{table:confusion-matrix}
        
        \end{table*}
        
        The metrics we use are defined as follows:
        \begin{itemize}
            \item {\bf Accuracy (A)}: fraction of true prediction over the total.
            \begin{eqnarray}
                \label{eq:accuracy} A=(TP+TN)/(TP+TN+FP+FN))
            \end{eqnarray} 
            
            \item {\bf F\textsubscript{1}-score ($F_{1}$)}: harmonic mean of Precision ($P$) and Recall ($R$), where $P$ is the fraction of relevant instances among the retrieved instances and $R$ represents the fraction of relevant instances that were retrieved:
            \begin{eqnarray}
                \label{eq:f1-score} F_{1}=2TP/(2TP+FP+FN)
            \end{eqnarray} 
            
            \item {\bf Jaccard (J)}: fraction of correctly predicted instances over all instances except those where a label is not present and is not predicted.
            \begin{eqnarray}
                \label{eq:jaccard} J=(TP)/(TP+FP+FN))
            \end{eqnarray} 
        \end{itemize} 
        To ensure all bias classes are treated equally, we use the macro-averaged $F_{1}$ and macro-averaged $J$ ($J_{m}$) scores to evaluate the overall performance of the models. 
        
        To evaluate the performance of the models for each language, we use the micro-averaged $J$ ($J_{\mu}$) score which accounts for the problem of class imbalance. Inspired by Nagle \citep{nagle1998proposal}, we also report the Relative Performance (RP) of the models for each language used in our study. RP is defined as the ratio of the absolute performance of the models under consideration. To compute it, any underlying evaluation metric (ex. $J$, $A$, $F_1$, etc.) can be used. In particular, we report on the relative performance of models trained with only headlines to those trained with or without additional knowledge and attention mechanism.

\section{Results and discussion}
\label{sec:results-and-discussion}

    We begin this section by analysing the experimental results of the models trained across all reported languages. Following that, we examine the performance of the models evaluated for individual languages. Finally, we present the findings of a case study that investigates the effect of translation quality on prediction accuracy.
    
    \subsection{Overall performance}
    \label{sec:overall-performance}
    
        We evaluate the baseline models and our proposed framework across all the reported languages and present their performance in terms of accuracy($A$), macro-averaged-F\textsubscript{1} ($F_{1}$), and macro-averaged-Jaccard($J_{m}$) scores in Table \ref{table:overall-performance-baselines}. As the results indicate, with $0.92$ $A$ and $F_{1}$, and $0.86$ $J_{m}$, our proposed framework trained with headlines and attended IC\_Knwl using cmlm-ml clearly outperforms other models trained with headlines only. It surpasses the performance of the best model (cmlm-ml) in terms of $A$ and $F_{1}$, and $J_{m}$ by $2.2\%$ and $3.6\%$ respectively.
    
        \begin{table*}[htb]
            \centering
            \small
            
            \begin{tabular}{lcccccc}
                 
                 & \textbf{ml-MiniLM} & \textbf{distil-mUSE} & \textbf{ml-mpnet} & \textbf{LaBSE} & \textbf{cmlm-ml} & \textbf{ours} \\ \hline
                
                \textbf{$\bm{A}$} & 0.62 & 0.64 & 0.66 & 0.75 & 0.90 & \textbf{0.92}\\ 
                \textbf{$\bm{F_{1}}$} & 0.57 & 0.61 & 0.63 & 0.74 & 0.90 & \textbf{0.92}\\ 
                
                \textbf{$\bm{J_{m}}$} & 0.40 & 0.44 & 0.46 & 0.59 & 0.83 & \textbf{0.86}\\ \hline 
                
            \end{tabular}
            
            \caption{Comparison between the baseline models and our proposed framework in terms of Accuracy($A$), macro-averaged-F\textsubscript{1} ($F_{1}$), and macro-averaged-Jaccard($J_{m}$) scores across all the reported languages. Trained with headlines and attended IC\_Knwl using cmlm-ml,  our framework outperforms the baseline models trained with headlines only.}
            
            \label{table:overall-performance-baselines}
            
        \end{table*}

        To determine whether the IC\_Knwl contains knowledge useful for bias prediction, we train the models with IC\_Knwl as the only input. We report the results in the first column of Table \ref{table:overall-performance}. We observe that models trained exclusively with IC\_Knwl achieve comparable results to models trained only with headlines. In terms of $A$, $F_{1}$, and $J_{m}$ scores, models other than cmlm-ml show an average improvement of $22\%$, $28\%$, and $47\%$, whereas cmlm-ml shows a slight decrease in performance of $5\%$, $2\%$, and $5\%$ respectively. The findings demonstrate that the IC\_Knwl does provide useful inferential information for the task of bias prediction (\textbf{RO2}).

        \begin{table*}[htb]
            \centering
            \small
            
            \begin{tabular}{l ccc ccc ccc}
                 & \multicolumn{3}{c}{\textbf{\begin{tabular}[c]{@{}c@{}}\\IC\_Knwl\end{tabular}}} &
                 \multicolumn{3}{c}{\textbf{\begin{tabular}[c]{@{}c@{}}Headline+\\IC\_Knwl\end{tabular}}} &
                 \multicolumn{3}{c}{\textbf{\begin{tabular}[c]{@{}c@{}}Headline+\\Attn(IC\_Knwl)\end{tabular}}} \\ \cline{2-10}
                 
                 & \textbf{$\bm{A}$} & \textbf{$\bm{F_{1}}$} & \textbf{$\bm{J_{m}}$} &
                 \textbf{$\bm{A}$} & \textbf{$\bm{F_{1}}$} & \textbf{$\bm{J_{m}}$} &
                 \textbf{$\bm{A}$} & \textbf{$\bm{F_{1}}$} & \textbf{$\bm{J_{m}}$} \\ \hline
                
                \textbf{ml-MiniLM} & 0.78 & 0.78 & 0.64 & 0.81 & 0.81 & 0.68 & 0.86 & 0.87 & 0.77 \smallskip \\ 
                
                \textbf{distil-mUSE} & 0.78 & 0.78 & 0.64 & 0.83 & 0.83 & 0.71 & 0.90 & 0.90 & 0.83 \smallskip \\ 
                
                \textbf{ml-mpnet} & 0.81 & 0.81 & 0.69 & 0.83 & 0.84 & 0.72 & 0.89 & 0.89 & 0.81 \smallskip \\ 
                
                \textbf{LaBSE} & 0.86 & 0.87 & 0.77 & 0.89 & 0.90 & 0.82 & 0.90 & 0.91 & 0.83 \smallskip \\ 
                
                \textbf{cmlm-ml} & 0.86 & 0.88 & 0.79 & 0.91 & 0.92 & 0.85 & 0.92 & 0.92 & 0.86\\ \hline
                
            \end{tabular}
            
            \caption{Accuracy($A$), macro-averaged-F\textsubscript{1} ($F_{1}$), and macro-averaged-Jaccard($J_{m}$) scores of the analysed models for all the reported languages. Each model is trained using IC\_Knwl, headlines with IC\_Knwl (Headline+IC\_Knwl), and headlines with attended IC\_Knwl (Headline+Attn(IC\_Knwl)) respectively.}
            \label{table:overall-performance}
            
        \end{table*}

        Furthermore, as evident in column two of Table \ref{table:overall-performance}, integrating IC\_Knwl with the headline can significantly improve the performance of all models by enhancing their reasoning abilities. In terms of $A$, $F_{1}$, and $J_{m}$ scores, these models exhibit average performance improvements of $4\%$, $4\%$, and $7\%$, respectively, over models trained exclusively with IC\_Knwl.
        
        Integration of IC\_Knwl, on the other hand, may not always function as expected and may introduce unwanted noises. Given the fact that they are generated automatically rather than manually, noise is inevitable, which may weaken their role in bias prediction. To minimise the impact of this noise, we integrate IC\_Knwl with an attention mechanism and present the results in column three of Table \ref{table:overall-performance}. The introduction of attention results in an average performance gain of $5\%$, $4\%$, and $9\%$ in terms of $A$, $F_{1}$, and $J_{m}$ scores, respectively.
        
        The good performance of the models can be attributed to their deep network architectures, which enable them to learn rich universal text representations. Furthermore, it demonstrates that integrating IC\_Knwl significantly improves their performance, while the introduction of attention improves it even further (\textbf{RO4}). To summarise, the results indicate that our proposed framework for bias prediction is effective regardless of the models used (\textbf{RO3}).  
        
    \subsection{Language-wise performance}
    \label{sec:language-wise-performance}
    
        The models evaluated for individual languages present plausible results, as shown in Table \ref{table:language-wise-performance}. However, the performance of models across languages varies significantly due to an imbalanced number of samples per class.
    
        \begin{table*}[!htbp]
            \centering
            \small
            
            \renewcommand{\arraystretch}{1.1}
            \begin{adjustbox}{width=1.0\textwidth}
                \begin{tabular}{ll ccc ccc ccc}
                     & & \multicolumn{3}{c}{\textbf{Headline}} & 
                     \multicolumn{3}{c}{\textbf{Headline+IC\_Knwl}} &
                     \multicolumn{3}{c}{\textbf{\begin{tabular}[c]{@{}c@{}}Headline+\\Attn(IC\_Knwl)\end{tabular}}}
                     \\ \cline{3-11}
                     
                     & & 
                     \textbf{$\bm{A}$} & \textbf{$\bm{F_1{}_{\mu}}$} &  \textbf{$\bm{J_{\mu}}$} & 
                     \textbf{$\bm{A}$} & \textbf{$\bm{F_1{}_{\mu}}$} &  \textbf{$\bm{J_{\mu}}$} &
                     \textbf{$\bm{A}$} & \textbf{$\bm{F_1{}_{\mu}}$} &  \textbf{$\bm{J_{\mu}}$}\\ \hline
                    
                    & \textbf{ml-MiniLM} & 0.53 & 0.53 & 0.36 & 1.05 & 1.03 & 1.05 & 1.67 & 1.16 & 1.23 \\ 
                    
                    & \textbf{distil-mUSE} & 0.53 & 0.53 & 0.36 & 1.15 & 1.15 & 1.19 & 1.16 & 1.16 & 1.27 \\ 
                    
                    \textbf{Slovenian} & \textbf{ml-mpnet} & 0.55 & 0.54 & 0.37 & 1.05 & 1.07 & 1.10 & 1.12 & 1.10 & 1.14 \\ 
                    
                    & \textbf{LaBSE} & 0.56 & 0.55 & 0.38 & 1.19 & 1.21 & 1.31 & 1.02 & 1.02 & 1.06 \\
                    
                    & \textbf{cmlm-ml} & 0.54 & 0.70 & 0.71 & 1.01 & 1.01 & 1.01 & 1.02 & 1.04 & 1.05 \\ \hline

                    
                    & \textbf{ml-MiniLM} & 0.47 & 0.47 & 0.31 & 1.87 & 1.87 & 2.54 & 1.06 & 1.05 & 1.11 \\ 
                    
                    & \textbf{distil-mUSE} & 0.55 & 0.55 & 0.38 & 1.50 & 1.50 & 1.86 & 1.14 & 1.13 & 1.25 \\ 
                    
                    \textbf{Romanian} & \textbf{ml-mpnet} & 0.56 & 0.56 & 0.38 & 1.60 & 1.58 & 2.13 & 1.05 & 1.05 & 1.09 \\ 
                    
                    & \textbf{LaBSE} & 0.81 & 0.81 & 0.68 & 1.14 & 1.14 & 1.27 & 1.02 & 1.02 & 1.03 \\
                    
                    & \textbf{cmlm-ml} & 0.95 & 0.94 & 0.89 & 1.00 & 1.01 & 1.01 & 1.01 & 1.00 & 1.01 \\ \hline
                    
                     
                    & \textbf{ml-MiniLM} & 0.52 & 0.51 & 0.34 & 1.71 & 1.74 & 2.35 & 1.08 & 1.08 & 1.17 \\ 
                    
                    & \textbf{distil-mUSE} & 0.56 & 0.56 & 0.38 & 1.60 & 1.58 & 2.23 & 1.10 & 1.11 & 1.20 \\ 
                    
                    \textbf{Swedish} & \textbf{ml-mpnet} & 0.58 & 0.58 & 0.41 & 1.58 & 1.58 & 2.07 & 1.07 & 1.07 & 1.15 \\ 
                    
                    & \textbf{LaBSE} & 0.78 & 0.78 & 0.64 & 1.26 & 1.26 & 1.54 & 1.00 & 1.00 & 1.00 \\
                    
                    & \textbf{cmlm-ml} & 0.98 & 0.98 & 0.96 & 1.01 & 1.01 & 1.03 & 1.00 & 1.00 & 1.00 \\   \hline
                    
                    
                    & \textbf{ml-MiniLM} & 0.81 & 0.81 & 0.68 & 1.17 & 1.17 & 1.33 & 1.00 & 1.00 & 1.00 \\ 
                    
                    & \textbf{distil-mUSE} & 0.79 & 0.78 & 0.64 & 1.24 & 1.24 & 1.48 & 1.01 & 1.02 & 1.03 \\
                    
                    \textbf{Finnish} & \textbf{ml-mpnet} & 0.82 & 0.82 & 0.69 & 1.18 & 1.18 & 1.36 & 1.02 & 1.02 & 1.04 \\ 
                    
                    & \textbf{LaBSE} & 0.84 & 0.84 & 0.72 & 1.17 & 1.17 & 1.36 & 1.00 & 1.00 & 1.00 \\
                    
                    & \textbf{cmlm-ml} & 0.99 & 0.99 & 0.98 & 1.00 & 1.00 & 1.01 & 1.00 & 1.00 & 1.00 \\ \hline

                    
                    & \textbf{ml-MiniLM} & 0.80 & 0.80 & 0.67 & 1.17 & 1.16 & 1.31 & 1.01 & 1.02 & 1.02 \\ 
                    
                    & \textbf{distil-mUSE} & 0.87 & 0.86 & 0.76 & 1.10 & 1.11 & 1.21 & 1.02 & 1.02 & 1.04 \\
                    
                    \textbf{Czech} & \textbf{ml-mpnet} & 0.84 & 0.83 & 0.72 & 1.15 & 1.16 & 1.30 & 1.02 & 1.02 & 1.04 \\

                    & \textbf{LaBSE} & 0.92 & 0.91 & 0.84 & 1.07 & 1.08 & 1.17 & 1.00 & 1.00 & 1.00 \\
                    
                    & \textbf{cmlm-ml} & 0.99 & 0.98 & 0.97 & 1.00 & 1.01 & 1.02 & 1.00 & 1.00 & 1.00 \\  \hline
                    
                \end{tabular}
            \end{adjustbox}
            
            \caption{Accuracy($A$), micro-averaged-F\textsubscript{1}($F_{1}{}_{\mu}$), and micro-averaged-Jaccard($J_{\mu}$) scores of the analysed models for each language used in the study. Each model is trained using headlines, headlines with IC\_Knwl (Headline+IC\_Knwl), and headlines with attended IC\_Knwl (Headline+Attn(IC\_Knwl)) respectively. For Headline+IC\_Knwl, we report its relative performance to the models trained with headlines only. For Headline+Attn(IC\_Knwl), we report its relative performance to the models trained with headlines and IC\_Knwl.}
            \label{table:language-wise-performance}
            
        \end{table*}
        
        Among all the low-resource languages present in the dataset used for this study, the models analysed for Czech demonstrate the most impressive performance, with an average $A$, $F_{1}{}_{\mu}$, and $J_{\mu}$ of $0.88$, $0.87$, and $0.79$ respectively for the models trained with headlines only. Since it leaves little room for performance improvement, models trained with additional IC\_Knwl with/without attention contribute an average of only ~$1.01$ times more to the calculated scores. 
        
        Following that, we have the models analysed for Finnish with the next best average $A$, $F_{1}{}_{\mu}$, and $J_{\mu}$ of $0.85$, $0.84$, and $0.79$ respectively for the models trained with headlines only. With the additional IC\_Knwl, the average scores for $A$ and $F_{1}{}_{\mu}$ increase by $1.15$ times and $J_{\mu}$ by $1.30$ times. Nonetheless, the benefits of employing attention are negligible.
        
        The impressive performance of the languages Czech and Finnish can be attributed to the fact that all of their samples belong to the class `Left-Center.' Since all of their bias labels are from the same class, it is possible that the classifiers may end up modelling the language specifics and writing style of the outlet in addition to the bias embedded in the headlines.
        
        The models evaluated for Swedish and Romanian produce the next best results that are nearly identical to each other, differing only by a small margin. For Swedish, models trained with only headlines show an average $A$/$F_{1}{}_{\mu}$ of $0.68$ and $J_{\mu}$ of $0.54$. These score differs by only $0.02$ points for Romanian. IC\_Knwl provides a substantial performance boost for both the languages. Swedish and Romanian have $A$/$F_{1}{}_{\mu}$ boosts of $1.43$ and $1.42$ times, and $J_{\mu}$ boosts of $1.84$ and $1.76$ times, respectively. They clearly benefit from the attention as well. Both the languages exhibit a $1.05$ times boost in $A$/$F_{1}{}_{\mu}$ and a ~$1.09$ times boost in $J_{\mu}$.
        
        In the case of the models analysed for Slovenian, one can notice a significant performance gap when compared to others. It demonstrates the lowest performance with an average $A$, $F_{1}{}_{\mu}$, and $J_{\mu}$ of $0.54$, $0.57$, and $0.43$ respectively for the PLMs trained with headlines only. With the additional IC\_Knwl, the average scores for $A$/$F_{1}{}_{\mu}$ increase by $1.09$ times and $J_{\mu}$ by $1.13$ times. Moreover, the benefits of employing attention can be noticed by a performance increase of $1.19$, $1.09$, and $1.15$ times in terms of $A$, $F_{1}{}_{\mu}$, and $J_{\mu}$ scores. Even with the highest number of examples ($~30\%$ ref. Table \ref{table:dataset-statistics}), its performance is low. To some extent, this could be attributed to the lower average headline length and language complexities that hinder the models' ability to comprehend the text for the task of bias prediction. Alternately, it could be due to the limited embedding coverage of the Slovenian language.
        
        Models trained with IC\_Knwl indicate low $A$ and high $J_{\mu}$. This implies that with more true negatives than true positives, performance evaluation using $A$ as the evaluation metric can turn out to be misleading. For instance, since there are no Slovenian examples that exist for the Right Center class, considering instances where they are not present and are not predicted (true negatives), would not be credible. In such cases, considering $J_{\mu}$ is more reliable since it disregards true negatives in favour of true positives.
        
        To sum up, across all languages analysed, cmlm-ml performed the best among models, whereas ml-MiniLM performed the worst. Overall, the results indicate the models trained with only headlines are capable of predicting bias inherent in them, even for low-resource languages like the ones used in this study. Moreover, IC\_Knwl significantly enhances model performance, especially when attention is employed.
        
    \subsection{Qualitative analysis}
    \label{sec:qualitative-analysis}
    
        In this section, we assess the effect of translation quality on prediction performance by analysing translation errors. We use the Slovenian language as a case study since the models analysed for it exhibit a significant performance gap when compared to other languages in our dataset. With the help of native Slovenian speakers in our research group, we discover several translation errors which we classify as follows:
        
        \begin{enumerate}
            \item {\textbf{Entity Detection Error:}} occurs when the translation engine misinterprets the entities referenced in the headline.
            \item {\textbf{Comprehension Error:} arises when the translation engine fails to comprehend the meaning of a headline, resulting in an unintelligible translation.}
            \item {\textbf{Improper Sentence Formation:}} when the translated headline grasps the basic idea of the original headline, but fails to form a coherent translated sentence, this error type occurs.
            \item {\textbf{Inversion of Meaning:}} takes place when the translation engine inverts the semantic meaning of a headline, resulting in a seemingly meaningful translation with a dissimilar semantic meaning.
            \item {\textbf{Miscellaneous Error:}} a category reserved for errors that do not fit into any of the aforementioned categories.
        \end{enumerate}
        
        \begin{table*}[!htbp]
        \centering
        \large
        
            \begin{adjustbox}{width=1.0\textwidth}
                
                \begin{tabular}{l m{1.0\textwidth}}
                
                \hline
                \multicolumn{2}{l}{\textbf{Entity Detection Error}} \\ \hline 
                
                \textbf{Slovenian Headline:} & \textcolor{Blue}{\textbf{Vodomec na 32. Liffu filmu Pohodi plin!}} \\ \hdashline[0.4pt/2pt] 
                
                \textbf{Generated Translation:} & Aquarius on the 32nd Liff movie Walk the Gas! \\ \hdashline[0.4pt/2pt] 
                
                \textbf{Correct Translation:} & \textcolor{Blue}{Kingfisher on the 32nd Liff awarded to movie Walk the Gas!} \\ \hdashline[0.4pt/2pt] 
                
                \textbf{Comment:} & The entity `\textbf{\textit{Vodomec}}', which means `\textbf{\textit{Common Kingfisher}}', is translated incorrectly as `\textbf{\textit{Aquarius}}'. However, it refers to the name of an award in this context. \\ \hline 
                
                \hline
                \multicolumn{2}{l}{\textbf{Comprehension Error}} \\ \hline 
                
                \textbf{Slovenian Headline:} & \textcolor{Blue}{\textbf{Počivalšek: Janša SMC ni ničesar prepustil}} \\ \hdashline[0.4pt/2pt] 
                
                \textbf{Generated Translation:} & Resting place: Janša SMC did not leave anything \\ \hdashline[0.4pt/2pt] 
                
                \textbf{Correct Translation:} & \textcolor{Blue}{Počivalšek: Janša left nothing for SMC} \\ \hdashline[0.4pt/2pt] 
                
                \textbf{Comment:} & The surname `\textbf{\textit{Počivalšek}}' is mistranslated as `\textbf{\textit{Resting place}}'. Furthermore, there exists no distinction between the surname `\textbf{\textit{Janša}}' and the political party `\textbf{\textit{SMC}}'. \\ \hline 
                
                \hline
                \multicolumn{2}{l}{\textbf{Improper Sentence Formation}} \\ \hline 
                
                \textbf{Slovenian Headline:} & \textcolor{Blue}{\textbf{Nad zdravstvene delavce z grožnjami in žalitvami}} \\ \hdashline[0.4pt/2pt] 
                
                \textbf{Generated Translation:} & Above health professionals with threats and insults \\ \hdashline[0.4pt/2pt] 
                
                \textbf{Correct Translation:} & \textcolor{Blue}{Threats, insults towards health professionals} \\ \hdashline[0.4pt/2pt] 
                
                \textbf{Comment:} & Depending on the context, `\textbf{\textit{Nad}}' could mean `\textbf{\textit{Above}}' or `\textbf{\textit{Towards}}'. The translation engine misinterprets `\textbf{\textit{Nad}}' in this case, resulting in an improper sentence formation. \\ \hline 
                
                \hline
                \multicolumn{2}{l}{\textbf{Inversion of Meaning}} \\ \hline 
                
                \textbf{Slovenian Headline:} & \textcolor{Blue}{\textbf{Na spletu podatki 533 milijonov Facebook uporabnikov, tudi 230.000 Slovencev}} \\ \hdashline[0.4pt/2pt] 
                
                \textbf{Generated Translation:} & There are 533 million Facebook users online, including 230,000 Slovenians \\ \hdashline[0.4pt/2pt] 
                
                \textbf{Correct Translation:} & \textcolor{Blue}{Data of 533 million Facebook users leaked online, including 230,000 Slovenians} \\ \hdashline[0.4pt/2pt] 
                
                \textbf{Comment:} & Although the translation is comprehensible, it refers to Facebook users instead of Facebook user data. \\ \hline 
                
                \hline
                \multicolumn{2}{l}{\textbf{Miscellaneous Error}} \\ \hline 
                
                \textbf{Slovenian Headline:} & \textcolor{Blue}{\textbf{Grujović naj bi streljal v silobranu, priča trdi drugače}} \\ \hdashline[0.4pt/2pt] 
                
                \textbf{Generated Translation:} & Grujović allegedly shot in the silobran, the witness claims otherwise \\ \hdashline[0.4pt/2pt] 
                
                \textbf{Correct Translation:} & \textcolor{Blue}{Grujović allegedly shot in self-defense, the witness claims otherwise} \\ \hdashline[0.4pt/2pt] 
                
                \textbf{Comment:} & Since `\textbf{\textit{silobranu}}' is misinterpreted as an entity, there is no attempt to translate `\textbf{\textit{v silobranu}}', which means `\textbf{\textit{in self-defense}}'. \\ \hline 
                
                \end{tabular}
            \end{adjustbox}
        \caption{Case study of Slovenian headlines to understand the translation error types (translation: from Slovenian to English).}
        \label{table:case-study}
        
        \end{table*}
        
        Table \ref{table:case-study} provides an example with appropriate justifications for each of these error types. In the majority of cases, the translation engine's lack of contextual awareness resulted in mistranslations. In some cases, the missing context could be inferred from the headline alone, whereas in others, reading the entire article or researching the entities mentioned in the headline appears to be the only way to obtain adequate context. Errors linked to a lack of vocabulary or other factors were less common. 
        
        Overall, the performance gap between Slovenian and other languages could be attributed to the language's poor translation quality relative to the other languages, as evidenced by the relatively numerous instances of improper translation. Given the complexity of the language and the small number of native speakers, the conclusion seems plausible.
        
\section{Research implications}
\label{sec:research-implications}   
    
    Predicting the political polarity of news headlines has many positive implications. It can not only help readers identify politically biased news but also allow journalists and the individuals involved in the news production process to assess their work objectively. Furthermore, such insights would also be interesting for researchers and social scientists. In this section, we further discuss the theoretical implications  of our research and the ways in which our proposed framework can enhance practical applications.
    
    \subsection{Theoretical implications}
    \label{sec:theoretical-implications}   
    
        Our study proposes a new perspective by leveraging Inferential Commonsense Knowledge (IC\_Knwl) via a Translate-Retrieve-Translate strategy to facilitate comprehension of the overall narrative of the multilingual headlines. Using IC\_Knwl, it introduces a language-agnostic learning framework to enhance the prediction of political polarity in multilingual news headlines. To the best of our knowledge, our proposed framework is one of the earliest attempts to leverage IC\_Knwl in a multilingual context for polarity prediction of news headlines. Since the existing work lacks annotated datasets for the task, it presents a dataset of multilingual news headlines. It simulates the real-world challenges of imbalanced data distribution by annotating headlines in five European low-resource languages with their respective political polarities. Our experimental investigation demonstrates the advantages of using IC\_Knwl, shedding light on the prospects of utilising it for the downstream tasks. It also demonstrates the effectiveness of multiple state-of-the-art multilingual pre-trained language models.
    
    \subsection{Practical implications}
    \label{sec:practical-implications}   

        Our study highlights the role of Inferential Commonsense Knowledge (IC\_Knwl) in facilitating the comprehension of short news headline text. It demonstrates that the IC\_Knwl, when used in conjunction with the translate-retrieve-translate technique, can effectively aid in the comprehension of narratives in a multilingual context. When fused with multilingual pre-trained language models (PLMs), it enhances the political polarity prediction of multilingual news headlines. Both implicit and explicit knowledge are expected in effective systems. The performance enhancement achieved by fusing the implicit knowledge obtained from the PLMs with explicit knowledge in the form of IC Knwl supports this view. 

        Given that a system is expected to deal with low-resource situations in the real world, our proposed framework is language-agnostic and thus adaptable to such scenarios. Another common problem in real world scenarios is scarcity of annotated data. Our proposed dataset, which focuses on low-resource languages with an imbalanced distribution, addresses this issue. Furthermore, our framework for data generation facilitates future expansion and the creation of custom datasets for related tasks.
        
\section{Conclusions and future works}
\label{sec:conclusions-and-future-works}

    In this paper, we introduced a language-agnostic learning framework infused with Inferential Commonsense Knowledge (IC\_Knwl) for enhancing the prediction of political polarity in multilingual news headlines under imbalanced sample distribution. We proposed to leverage IC\_Knwl through a Translate-Retrieve-Translate (TRT) strategy to help uncover contextual features for comprehension of the overall narrative of the multilingual headlines. Since not all the retrieved inferences are expected to be of equal relevance, we also employed an attention mechanism to emphasise relevant inferences. We used the neural-network model COMET trained on the $\text{ATOMIC}^{20}_{20}$ knowledge graphs to retrieve IC\_Knwl and employed the Google Translate API for translation. Furthermore, we presented an annotated dataset of news headlines in five low-resource European languages.

    We conducted an extensive evaluation of our framework with several multilingual pre-trained language models (PLMs). The evaluation results revealed their impressive performance, which can be attributed to their complex network architectures. The results also demonstrated that incorporating IC\_Knwl and employing attention significantly enhanced their performance. Overall, the results indicate that the proposed framework for bias prediction is effective regardless of the models used. Even the models evaluated for individual languages present plausible results. Furthermore, we conducted a thorough case study on the Slovenian headlines to investigate translation errors. The study uncovered numerous instances of improper translation, indicating that the performance gap between Slovenian and other languages may be attributable to the language's poor translation quality.
    
    In the future, we plan to diversify our additional knowledge sources. In particular, we intend to investigate how knowledge sources such as \href{https://www.wiktionary.org}{Wiktionary} and \href{https://conceptnet.io/}{ConceptNet} influence the task of polarity prediction. Another possible direction is to extend this study beyond polarity prediction to its quantification and correction. It would also be interesting to experiment with auxiliary tasks involving news headlines in a multitask learning paradigm. 

\section*{CRediT authorship contribution statement}
    
    \textbf{Swati Swati:} Conceptualization, Data curation, Investigation, Methodology, Software, Validation, Visualization, Writing - original draft. \textbf{Adrian Mladenić Grobelnik:} Investigation, Validation, Writing - review \& editing. \textbf{Dunja Mladenić:} Conceptualization, Supervision, Writing - review \& editing. \textbf{Marko Grobelnik:} Conceptualization, Funding acquisition, Supervision, Writing - review \& editing.

\section*{Declaration of competing interest}
    The authors declare that they have no known competing financial interests or personal relationships that could have appeared to influence the work reported in this paper.

\section*{Data availability}
    Dataset and scripts available at: \textcolor{Blue}{\url{https://github.com/Swati17293/KG-Multi-Bias}}

\section*{Acknowledgements}
    This work was supported by the Slovenian Research Agency under the project J2-1736 Causalify and the European Union’s Horizon 2020 research and innovation program under the Marie Skłodowska-Curie grant agreement No 812997.

\bibliography{main}

\begin{thebibliography}{100}
\expandafter\ifx\csname url\endcsname\relax
  \def\url#1{\texttt{#1}}\fi
\expandafter\ifx\csname urlprefix\endcsname\relax\def\urlprefix{URL }\fi
\expandafter\ifx\csname href\endcsname\relax
  \def\href#1#2{#2} \def\path#1{#1}\fi

\bibitem{helberger2019democratic}
N.~Helberger, On the democratic role of news recommenders, Digital Journalism
  7~(8) (2019) 993--1012.
\newblock \href {http://dx.doi.org/10.1080/21670811.2019.1623700}
  {\path{doi:10.1080/21670811.2019.1623700}}.

\bibitem{mcnair2009journalism}
B.~McNair, Journalism and democracy, in: The handbook of journalism studies,
  Routledge, 2009, pp. 257--269.
\newblock \href {http://dx.doi.org/10.4324/9780203877685-27}
  {\path{doi:10.4324/9780203877685-27}}.

\bibitem{miller2000news}
J.~M. Miller, J.~A. Krosnick, News media impact on the ingredients of
  presidential evaluations: Politically knowledgeable citizens are guided by a
  trusted source, American Journal of Political Science (2000) 301--315\href
  {http://dx.doi.org/10.2307/2669312} {\path{doi:10.2307/2669312}}.

\bibitem{park2009newscube}
S.~Park, S.~Kang, S.~Chung, J.~Song, Newscube: delivering multiple aspects of
  news to mitigate media bias, in: Proceedings of the SIGCHI conference on
  human factors in computing systems, Association for Computing Machinery,
  2009, pp. 443--452.
\newblock \href {http://dx.doi.org/10.1145/1518701.1518772}
  {\path{doi:10.1145/1518701.1518772}}.

\bibitem{davis2022gender}
S.~R. Davis, C.~J. Worsnop, E.~M. Hand, Gender bias recognition in political
  news articles, Machine Learning with Applications 8 (2022) 100304.
\newblock \href {http://dx.doi.org/10.1016/j.mlwa.2022.100304}
  {\path{doi:10.1016/j.mlwa.2022.100304}}.

\bibitem{spinde2021interdisciplinary}
T.~Spinde, An interdisciplinary approach for the automated detection and
  visualization of media bias in news articles, in: 2021 International
  Conference on Data Mining Workshops (ICDMW), IEEE, 2021, pp. 1096--1103.
\newblock \href {http://dx.doi.org/10.1109/ICDMW53433.2021.00144}
  {\path{doi:10.1109/ICDMW53433.2021.00144}}.

\bibitem{spinde2021automated}
T.~Spinde, L.~Rudnitckaia, J.~Mitrovi{\'c}, F.~Hamborg, M.~Granitzer, B.~Gipp,
  K.~Donnay, Automated identification of bias inducing words in news articles
  using linguistic and context-oriented features, Information Processing \&
  Management 58~(3) (2021) 102505.
\newblock \href {http://dx.doi.org/10.1016/j.ipm.2021.102505}
  {\path{doi:10.1016/j.ipm.2021.102505}}.

\bibitem{chen2022partisan}
K.~Chen, M.~Babaeianjelodar, Y.~Shi, K.~Janmohamed, R.~Sarkar, I.~Weber,
  T.~Davidson, M.~De~Choudhury, S.~Yadav, A.~Khudabukhsh, et~al., Partisan us
  news media representations of syrian refugees, arXiv preprint
  arXiv:2206.09024\href {http://dx.doi.org/10.48550/arXiv.2206.09024}
  {\path{doi:10.48550/arXiv.2206.09024}}.

\bibitem{chipidza2021effect}
W.~Chipidza, The effect of toxicity on covid-19 news network formation in
  political subcommunities on reddit: An affiliation network approach,
  International Journal of Information Management 61 (2021) 102397.
\newblock \href {http://dx.doi.org/10.1016/j.ijinfomgt.2021.102397}
  {\path{doi:10.1016/j.ijinfomgt.2021.102397}}.

\bibitem{hamborg2019automated}
F.~Hamborg, K.~Donnay, B.~Gipp, Automated identification of media bias in news
  articles: an interdisciplinary literature review, International Journal on
  Digital Libraries 20~(4) (2019) 391--415.
\newblock \href {http://dx.doi.org/https://doi.org/10.1007/s00799-018-0261-y}
  {\path{doi:https://doi.org/10.1007/s00799-018-0261-y}}.

\bibitem{gangula2019detecting}
R.~R.~R. Gangula, S.~R. Duggenpudi, R.~Mamidi, Detecting political bias in news
  articles using headline attention, in: Proceedings of the 2019 ACL Workshop
  BlackboxNLP: Analyzing and Interpreting Neural Networks for NLP, Association
  for Computational Linguistics, 2019, pp. 77--84.
\newblock \href {http://dx.doi.org/10.18653/v1/W19-4809}
  {\path{doi:10.18653/v1/W19-4809}}.

\bibitem{laban2021news}
P.~Laban, L.~Bandarkar, M.~A. Hearst, News headline grouping as a challenging
  nlu task, in: Proceedings of the 2021 Conference of the North American
  Chapter of the Association for Computational Linguistics: Human Language
  Technologies, Association for Computational Linguistics, 2021, pp.
  3186--3198.
\newblock \href {http://dx.doi.org/10.18653/v1/2021.naacl-main.255}
  {\path{doi:10.18653/v1/2021.naacl-main.255}}.

\bibitem{holmqvist2003reading}
K.~Holmqvist, J.~Holsanova, M.~Barthelson, D.~Lundqvist, Reading or scanning? a
  study of newspaper and net paper reading, in: The Mind's Eye, Elsevier, 2003,
  pp. 657--670.
\newblock \href {http://dx.doi.org/10.1016/B978-044451020-4/50035-9}
  {\path{doi:10.1016/B978-044451020-4/50035-9}}.

\bibitem{andrew2007media}
B.~C. Andrew, Media-generated shortcuts: Do newspaper headlines present another
  roadblock for low-information rationality?, Harvard International Journal of
  Press/Politics 12~(2) (2007) 24--43.
\newblock \href {http://dx.doi.org/10.1177/1081180X07299795}
  {\path{doi:10.1177/1081180X07299795}}.

\bibitem{ecker2014effects}
U.~K. Ecker, S.~Lewandowsky, E.~P. Chang, R.~Pillai, The effects of subtle
  misinformation in news headlines., Journal of experimental psychology:
  applied 20~(4) (2014) 323.
\newblock \href {http://dx.doi.org/10.1037/xap0000028}
  {\path{doi:10.1037/xap0000028}}.

\bibitem{molek2013towards}
K.~Molek-Kozakowska, Towards a pragma-linguistic framework for the study of
  sensationalism in news headlines, Discourse \& Communication 7~(2) (2013)
  173--197.
\newblock \href {http://dx.doi.org/10.1177/1750481312471668}
  {\path{doi:10.1177/1750481312471668}}.

\bibitem{ifantidou2009newspaper}
E.~Ifantidou, Newspaper headlines and relevance: Ad hoc concepts in ad hoc
  contexts, Journal of Pragmatics 41~(4) (2009) 699--720.
\newblock \href {http://dx.doi.org/10.1016/j.pragma.2008.10.016}
  {\path{doi:10.1016/j.pragma.2008.10.016}}.

\bibitem{mccluskey2005content}
M.~McCluskey, A content analysis of 2004 presidential election headlines of the
  los angeles times and the washington times, Electronic Theses and
  Dissertations 358, \url{https://stars.library.ucf.edu/etd/358}.

\bibitem{jovanovic2021headlines}
S.~M. Jovanovi{\'c}, et~al., Headlines against democracy: Operational code
  analysis of the serbian daily informer’s headlines in relation to the
  anti-government protests’ first phase (2018--2019), Journal of Media
  Research-Revista de Studii Media 14~(41) (2021) 23--41,
  \url{https://www.ceeol.com/search/article-detail?id=1000890}.

\bibitem{zeng2018critical}
D.~Zeng, W.~Gong, S.~Li, Critical discourse analysis on the news headline about
  terrorism: A case study of the english reports on counter-terrorism from
  turkey, Modern Linguistics 6~(3) (2018) 496--501.
\newblock \href {http://dx.doi.org/10.12677/ml.2018.63057}
  {\path{doi:10.12677/ml.2018.63057}}.

\bibitem{andrew2013political}
B.~Andrew, Political journalism represented by headline news: Canadian public
  and commercial media compared, Canadian Journal of Political Science/Revue
  canadienne de science politique 46~(2) (2013) 455--478.
\newblock \href {http://dx.doi.org/10.1017/S0008423913000462}
  {\path{doi:10.1017/S0008423913000462}}.

\bibitem{navia2015mercurio}
P.~Navia, R.~Osorio, El mercurio lies, and la tercera lies more. political bias
  in newspaper headlines in chile, 1994--2010, Bulletin of Latin American
  Research 34~(4) (2015) 467--485.
\newblock \href {http://dx.doi.org/10.1111/blar.12364}
  {\path{doi:10.1111/blar.12364}}.

\bibitem{hamborg2017identification}
F.~Hamborg, N.~Meuschke, A.~Aizawa, B.~Gipp, Identification and analysis of
  media bias in news articles, 2017.
\newblock \href {http://dx.doi.org/10.18452/1446} {\path{doi:10.18452/1446}}.

\bibitem{hamborg2020bias}
F.~Hamborg, N.~Meuschke, B.~Gipp, Bias-aware news analysis using matrix-based
  news aggregation, International Journal on Digital Libraries 21~(2) (2020)
  129--147.
\newblock \href {http://dx.doi.org/10.1007/s00799-018-0239-9}
  {\path{doi:10.1007/s00799-018-0239-9}}.

\bibitem{aksenov2021fine}
D.~Aksenov, P.~Bourgonje, K.~Zaczynska, M.~Ostendorff, J.~M. Schneider,
  G.~Rehm, Fine-grained classification of political bias in german news: A data
  set and initial experiments, in: Proceedings of the 5th Workshop on Online
  Abuse and Harms (WOAH 2021), 2021, pp. 121--131.
\newblock \href {http://dx.doi.org/10.18653/v1/2021.woah-1.13}
  {\path{doi:10.18653/v1/2021.woah-1.13}}.

\bibitem{guo2022modeling}
S.~Guo, K.~Q. Zhu, Modeling multi-level context for informational bias
  detection by contrastive learning and sentential graph network, arXiv
  preprint arXiv:2201.10376\href {http://dx.doi.org/10.48550/arXiv.2201.10376}
  {\path{doi:10.48550/arXiv.2201.10376}}.

\bibitem{doan2022survey}
T.~M. Doan, J.~A. Gulla, A survey on political viewpoints identification,
  Online Social Networks and Media 30 (2022) 100208.
\newblock \href {http://dx.doi.org/10.1016/j.osnem.2022.100208}
  {\path{doi:10.1016/j.osnem.2022.100208}}.

\bibitem{park2012computational}
S.~Park, S.~Kang, S.~Chung, J.~Song, A computational framework for media bias
  mitigation, ACM Transactions on Interactive Intelligent Systems (TiiS) 2~(2)
  (2012) 1--32.
\newblock \href {http://dx.doi.org/10.1145/2209310.2209311}
  {\path{doi:10.1145/2209310.2209311}}.

\bibitem{del2014lremap}
R.~Del~Gratta, F.~Frontini, A.~F. Khan, J.~Mariani, C.~Soria, The lremap for
  under-resourced languages, in: Workshop Collaboration and Computing for
  Under-Resourced Languages in the Linked Open Data Era, Satellite Workshop of
  LREC, Vol.~14, 2014,
  \url{https://www.academia.edu/download/39974165/The_LREMap_for_Under-Resourced_Languages20151113-28056-rmmjbf.pdf}.

\bibitem{bruneau2012going}
E.~G. Bruneau, M.~Cikara, R.~Saxe, Going beyond the headlines: Narratives
  mitigate intergroup empathy bias., in: Proceedings of the 34th Annual Meeting
  of the Cognitive Science Society, CogSci 2012, cognitivesciencesociety.org,
  2012, \url{https://mindmodeling.org/cogsci2012/papers/0475/index.html}.

\bibitem{berner1983commentary}
R.~T. Berner, Commentary: The narrative and the headline, Newspaper Research
  Journal 4~(3) (1983) 33--40.
\newblock \href {http://dx.doi.org/10.1177/073953298300400305}
  {\path{doi:10.1177/073953298300400305}}.

\bibitem{li2021enhancing}
D.~Li, X.~Zhu, Y.~Li, S.~Wang, D.~Li, J.~Liao, J.~Zheng, Enhancing emotion
  inference in conversations with commonsense knowledge, Knowledge-Based
  Systems 232 (2021) 107449.
\newblock \href {http://dx.doi.org/10.1016/j.knosys.2021.107449}
  {\path{doi:10.1016/j.knosys.2021.107449}}.

\bibitem{li2021past}
J.~Li, Z.~Lin, P.~Fu, W.~Wang, Past, present, and future: Conversational
  emotion recognition through structural modeling of psychological knowledge,
  in: Findings of the Association for Computational Linguistics: EMNLP 2021,
  Association for Computational Linguistics, 2021, pp. 1204--1214.
\newblock \href {http://dx.doi.org/10.18653/v1/2021.findings-emnlp.104}
  {\path{doi:10.18653/v1/2021.findings-emnlp.104}}.

\bibitem{du2022enhancing}
L.~Du, X.~Ding, K.~Xiong, T.~Liu, B.~Qin, Enhancing pretrained language models
  with structured commonsense knowledge for textual inference, Knowledge-Based
  Systems 254 (2022) 109488.
\newblock \href {http://dx.doi.org/10.1016/j.knosys.2022.109488}
  {\path{doi:10.1016/j.knosys.2022.109488}}.

\bibitem{li2021enhancing2}
R.~Li, Z.~Jiang, L.~Wang, X.~Lu, M.~Zhao, D.~Chen, Enhancing transformer-based
  language models with commonsense representations for knowledge-driven machine
  comprehension, Knowledge-Based Systems 220 (2021) 106936.
\newblock \href {http://dx.doi.org/10.1016/j.knosys.2021.106936}
  {\path{doi:10.1016/j.knosys.2021.106936}}.

\bibitem{lieto2021commonsense}
A.~Lieto, G.~L. Pozzato, S.~Zoia, V.~Patti, R.~Damiano, A commonsense reasoning
  framework for explanatory emotion attribution, generation and
  re-classification, Knowledge-Based Systems 227 (2021) 107166.
\newblock \href {http://dx.doi.org/10.1016/j.knosys.2021.107166}
  {\path{doi:10.1016/j.knosys.2021.107166}}.

\bibitem{swati4114271ic}
S.~Swati, M.~Grobelnik, Ic-bait: An inferential commonsense-driven model for
  predicting political polarity in news headlines, Available at SSRN
  4114271\href {http://dx.doi.org/10.2139/ssrn.4114271}
  {\path{doi:10.2139/ssrn.4114271}}.

\bibitem{hwang2021comet}
J.~D. Hwang, C.~Bhagavatula, R.~Le~Bras, J.~Da, K.~Sakaguchi, A.~Bosselut,
  Y.~Choi, (comet-) atomic 2020: On symbolic and neural commonsense knowledge
  graphs, in: Proceedings of the AAAI Conference on Artificial Intelligence,
  Vol.~35, 2021, pp. 6384--6392,
  \url{https://ojs.aaai.org/index.php/AAAI/article/view/16792}.

\bibitem{fang2022leveraging}
Y.~Fang, S.~Wang, Y.~Xu, R.~Xu, S.~Sun, C.~Zhu, M.~Zeng, Leveraging knowledge
  in multilingual commonsense reasoning, in: Findings of the Association for
  Computational Linguistics: ACL 2022, Association for Computational
  Linguistics, 2022, pp. 3237--3246.
\newblock \href {http://dx.doi.org/10.18653/v1/2022.findings-acl.255}
  {\path{doi:10.18653/v1/2022.findings-acl.255}}.

\bibitem{bonyadi2013headlines}
A.~Bonyadi, M.~Samuel, Headlines in newspaper editorials: A contrastive study,
  Sage Open 3~(2) (2013) 2158244013494863.
\newblock \href {http://dx.doi.org/10.1177/2158244013494863}
  {\path{doi:10.1177/2158244013494863}}.

\bibitem{MBFCFinnish}
MBFC, Finnish news - media bias/fact check -
  https://mediabiasfactcheck.com/finnish-news/,
  \url{https://mediabiasfactcheck.com/finnish-news/}, accessed: July 4, 2022.

\bibitem{li2022cross}
M.~Li, H.~Zhou, J.~Hou, P.~Wang, E.~Gao, Is cross-linguistic advert flaw
  detection in wikipedia feasible? a multilingual-bert-based transfer learning
  approach, Knowledge-Based Systems 252 (2022) 109330.
\newblock \href {http://dx.doi.org/10.1016/j.knosys.2022.109330}
  {\path{doi:10.1016/j.knosys.2022.109330}}.

\bibitem{lu2015transfer}
J.~Lu, V.~Behbood, P.~Hao, H.~Zuo, S.~Xue, G.~Zhang, Transfer learning using
  computational intelligence: A survey, Knowledge-Based Systems 80 (2015)
  14--23.
\newblock \href {http://dx.doi.org/10.1016/j.knosys.2015.01.010}
  {\path{doi:10.1016/j.knosys.2015.01.010}}.

\bibitem{pamungkas2021joint}
E.~W. Pamungkas, V.~Basile, V.~Patti, A joint learning approach with knowledge
  injection for zero-shot cross-lingual hate speech detection, Information
  Processing \& Management 58~(4) (2021) 102544.
\newblock \href {http://dx.doi.org/10.1016/j.ipm.2021.102544}
  {\path{doi:10.1016/j.ipm.2021.102544}}.

\bibitem{feng2020language}
F.~Feng, Y.~Yang, D.~Cer, N.~Arivazhagan, W.~Wang, Language-agnostic bert
  sentence embedding, arXiv preprint arXiv:2007.01852\href
  {http://dx.doi.org/10.48550/arXiv.2007.01852}
  {\path{doi:10.48550/arXiv.2007.01852}}.

\bibitem{reimers2019sentence}
N.~Reimers, I.~Gurevych, Sentence-bert: Sentence embeddings using siamese
  bert-networks, in: Proceedings of the 2019 Conference on Empirical Methods in
  Natural Language Processing and the 9th International Joint Conference on
  Natural Language Processing (EMNLP-IJCNLP), 2019, pp. 3982--3992.
\newblock \href {http://dx.doi.org/10.18653/v1/D19-1410}
  {\path{doi:10.18653/v1/D19-1410}}.

\bibitem{yang2021universal}
Z.~Yang, Y.~Yang, D.~Cer, J.~Law, E.~Darve, Universal sentence representation
  learning with conditional masked language model, in: Proceedings of the 2021
  Conference on Empirical Methods in Natural Language Processing, 2021, pp.
  6216--6228.
\newblock \href {http://dx.doi.org/10.18653/v1/2021.emnlp-main.502}
  {\path{doi:10.18653/v1/2021.emnlp-main.502}}.

\bibitem{kruspe2020cross}
A.~Kruspe, M.~H{\"a}berle, I.~Kuhn, X.~X. Zhu, Cross-language sentiment
  analysis of european twitter messages during the covid-19 pandemic, in:
  Proceedings of the 1st Workshop on {NLP} for COVID-19 at ACL 2020,
  Association for Computational Linguistics, 2020,
  \url{https://aclanthology.org/2020.nlpcovid19-acl.14}.

\bibitem{pei2022ab}
Y.~Pei, S.~Chen, Z.~Ke, W.~Silamu, Q.~Guo, Ab-labse: Uyghur sentiment analysis
  via the pre-training model with bilstm, Applied Sciences 12~(3) (2022) 1182.
\newblock \href {http://dx.doi.org/10.3390/app12031182}
  {\path{doi:10.3390/app12031182}}.

\bibitem{patel2021efficient}
R.~N. Patel, E.~Burgin, H.~Assem, S.~Dutta, Efficient multi-lingual sentence
  classification framework with sentence meta encoders, in: 2021 IEEE
  International Conference on Big Data (Big Data), IEEE, 2021, pp. 1889--1899.
\newblock \href {http://dx.doi.org/10.1109/BigData52589.2021.9671714}
  {\path{doi:10.1109/BigData52589.2021.9671714}}.

\bibitem{talat2022you}
Z.~Talat, A.~Neveol, S.~Biderman, M.~Clinciu, M.~Dey, S.~Longpre, S.~Luccioni,
  M.~Masoud, M.~Mitchell, D.~Radev, et~al., You reap what you sow: On the
  challenges of bias evaluation under multilingual settings, in: Proceedings of
  BigScience Episode\# 5--Workshop on Challenges \& Perspectives in Creating
  Large Language Models, Association for Computational Linguistics, 2022, pp.
  26--41.
\newblock \href {http://dx.doi.org/10.18653/v1/2022.bigscience-1.3}
  {\path{doi:10.18653/v1/2022.bigscience-1.3}}.

\bibitem{roy2019deep}
A.~Roy, K.~Basak, A.~Ekbal, P.~Bhattacharyya, A deep ensemble framework for
  fake news detection and multi-class classification of short political
  statements, in: Proceedings of the 16th International Conference on Natural
  Language Processing, 2019, pp. 9--17,
  \url{https://aclanthology.org/2019.icon-1.2/}.

\bibitem{saikh2019novel}
T.~Saikh, A.~Anand, A.~Ekbal, P.~Bhattacharyya, A novel approach towards fake
  news detection: deep learning augmented with textual entailment features, in:
  International Conference on Applications of Natural Language to Information
  Systems, Springer, 2019, pp. 345--358.
\newblock \href {http://dx.doi.org/10.1007/978-3-030-23281-8_30}
  {\path{doi:10.1007/978-3-030-23281-8_30}}.

\bibitem{saikh2019deep}
T.~Saikh, A.~De, A.~Ekbal, P.~Bhattacharyya, A deep learning approach for
  automatic detection of fake news, in: Proceedings of the 16th International
  Conference on Natural Language Processing, 2019, pp. 230--238.
\newblock \href {http://dx.doi.org/https://aclanthology.org/2019.icon-1.27}
  {\path{doi:https://aclanthology.org/2019.icon-1.27}}.

\bibitem{king2021diffusion}
K.~K. King, B.~Wang, Diffusion of real versus misinformation during a crisis
  event: a big data-driven approach, International Journal of Information
  Management (2021) 102390\href
  {http://dx.doi.org/10.1016/j.ijinfomgt.2021.102390}
  {\path{doi:10.1016/j.ijinfomgt.2021.102390}}.

\bibitem{rotim2017takelab}
L.~Rotim, M.~Tutek, J.~{\v{S}}najder, Takelab at semeval-2017 task 5: Linear
  aggregation of word embeddings for fine-grained sentiment analysis of
  financial news, in: Proceedings of the 11th International Workshop on
  Semantic Evaluation (SemEval-2017), 2017, pp. 866--871.
\newblock \href {http://dx.doi.org/10.18653/v1/S17-2148}
  {\path{doi:10.18653/v1/S17-2148}}.

\bibitem{korenvcic2018document}
D.~Koren{\v{c}}i{\'c}, S.~Ristov, J.~{\v{S}}najder, Document-based topic
  coherence measures for news media text, Expert systems with Applications 114
  (2018) 357--373.
\newblock \href {http://dx.doi.org/10.1016/j.eswa.2018.07.063}
  {\path{doi:10.1016/j.eswa.2018.07.063}}.

\bibitem{pandur2020topic}
M.~B. Pandur, J.~Dob{\v{s}}a, S.~Beliga, A.~Me{\v{s}}trovi{\'c}, Topic
  modelling and sentiment analysis of covid-19 related news on croatian
  internet portal, Information Society 2020 (2020) 5--9.

\bibitem{muller2021multimodal}
E.~M{\"u}ller-Budack, J.~Theiner, S.~Diering, M.~Idahl, S.~Hakimov, R.~Ewerth,
  Multimodal news analytics using measures of cross-modal entity and context
  consistency, International Journal of Multimedia Information Retrieval 10~(2)
  (2021) 111--125.
\newblock \href {http://dx.doi.org/https://doi.org/10.1007/s13735-021-00207-4}
  {\path{doi:https://doi.org/10.1007/s13735-021-00207-4}}.

\bibitem{tahmasebzadeh2020feature}
G.~Tahmasebzadeh, S.~Hakimov, E.~M{\"u}ller-Budack, R.~Ewerth, A feature
  analysis for multimodal news retrieval, in: Proceedings of the 1st
  International Workshop on Cross-lingual Event-centric Open Analytics
  co-located with the 17th Extended Semantic Web Conference (ESWC 2020),
  Aachen: RWTH, 2020.
\newblock \href {http://dx.doi.org/https://doi.org/10.34657/5197}
  {\path{doi:https://doi.org/10.34657/5197}}.

\bibitem{d2012media}
D.~D'Alessio, Media bias in presidential election coverage, 1948-2008:
  Evaluation via formal measurement, Lexington Books, 2012.

\bibitem{palic2019takelab}
N.~Pali{\'c}, J.~Vladika, D.~{\v{C}}ubeli{\'c}, I.~Lovren{\v{c}}i{\'c},
  M.~Buljan, J.~{\v{S}}najder, Takelab at semeval-2019 task 4: Hyperpartisan
  news detection, in: Proceedings of the 13th International Workshop on
  Semantic Evaluation, 2019, pp. 995--998.
\newblock \href {http://dx.doi.org/10.18653/v1/S19-2172}
  {\path{doi:10.18653/v1/S19-2172}}.

\bibitem{stevenson1980reconsideration}
R.~L. Stevenson, M.~T. Greene, A reconsideration of bias in the news,
  Journalism Quarterly 57~(1) (1980) 115--121.
\newblock \href {http://dx.doi.org/10.1177/107769908005700117}
  {\path{doi:10.1177/107769908005700117}}.

\bibitem{chen2018learning}
W.-F. Chen, H.~Wachsmuth, K.~Al~Khatib, B.~Stein, Learning to flip the bias of
  news headlines, in: Proceedings of the 11th International Conference on
  Natural Language Generation, Association for Computational Linguistics, 2018,
  pp. 79--88.
\newblock \href {http://dx.doi.org/10.18653/v1/W18-6509}
  {\path{doi:10.18653/v1/W18-6509}}.

\bibitem{groseclose2005measure}
T.~Groseclose, J.~Milyo, A measure of media bias, The Quarterly Journal of
  Economics 120~(4) (2005) 1191--1237.
\newblock \href {http://dx.doi.org/10.1162/003355305775097542}
  {\path{doi:10.1162/003355305775097542}}.

\bibitem{iyyer2014political}
M.~Iyyer, P.~Enns, J.~Boyd-Graber, P.~Resnik, Political ideology detection
  using recursive neural networks, in: Proceedings of the 52nd Annual Meeting
  of the Association for Computational Linguistics (Volume 1: Long Papers),
  2014, pp. 1113--1122, \url{https://aclanthology.org/P14-1105.pdf}.

\bibitem{naredla2022detection}
N.~R. Naredla, F.~F. Adedoyin, Detection of hyperpartisan news articles using
  natural language processing technique, International Journal of Information
  Management Data Insights 2~(1) (2022) 100064.
\newblock \href {http://dx.doi.org/10.1016/j.jjimei.2022.100064}
  {\path{doi:10.1016/j.jjimei.2022.100064}}.

\bibitem{tourni2021detecting}
I.~Tourni, L.~Guo, T.~H. Daryanto, F.~Zhafransyah, E.~E. Halim, M.~Jalal,
  B.~Chen, S.~Lai, H.~Hu, M.~Betke, et~al., Detecting frames in news headlines
  and lead images in us gun violence coverage, in: Findings of the Association
  for Computational Linguistics: EMNLP 2021, 2021, pp. 4037--4050.
\newblock \href {http://dx.doi.org/10.18653/v1/2021.findings-emnlp.339}
  {\path{doi:10.18653/v1/2021.findings-emnlp.339}}.

\bibitem{krieger2022domain}
J.-D. Krieger, T.~Spinde, T.~Ruas, J.~Kulshrestha, B.~Gipp, A domain-adaptive
  pre-training approach for language bias detection in news, in: Proceedings of
  the 22nd ACM/IEEE Joint Conference on Digital Libraries, 2022, pp. 1--7.
\newblock \href {http://dx.doi.org/10.1145/3529372.3530932}
  {\path{doi:10.1145/3529372.3530932}}.

\bibitem{magotra2022news}
V.~Magotra, E.~Hirani, V.~Mehta, S.~Dholay, News bias detection using
  transformers, in: Communication and Intelligent Systems, Springer, 2022, pp.
  319--326.
\newblock \href {http://dx.doi.org/10.1007/978-981-19-2130-8_26}
  {\path{doi:10.1007/978-981-19-2130-8_26}}.

\bibitem{hoyer2016spanish}
A.~Hoyer, Spanish news framing of the syrian refugee crisis, WWU Honors Program
  Senior Projects 26, \url{https://cedar.wwu.edu/wwu_honors/26/}.

\bibitem{fan2019plain}
L.~Fan, M.~White, E.~Sharma, R.~Su, P.~K. Choubey, R.~Huang, L.~Wang, In plain
  sight: Media bias through the lens of factual reporting, in: Proceedings of
  the 2019 Conference on Empirical Methods in Natural Language Processing and
  the 9th International Joint Conference on Natural Language Processing
  (EMNLP-IJCNLP), 2019, pp. 6343--6349.
\newblock \href {http://dx.doi.org/10.18653/v1/D19-1664}
  {\path{doi:10.18653/v1/D19-1664}}.

\bibitem{baly2020we}
R.~Baly, G.~Da~San~Martino, J.~Glass, P.~Nakov, We can detect your bias:
  Predicting the political ideology of news articles, in: Proceedings of the
  2020 Conference on Empirical Methods in Natural Language Processing (EMNLP),
  Association for Computational Linguistics, 2020, pp. 4982--4991.
\newblock \href {http://dx.doi.org/10.18653/v1/2020.emnlp-main.404}
  {\path{doi:10.18653/v1/2020.emnlp-main.404}}.

\bibitem{petroni2019language}
F.~Petroni, T.~Rockt{\"a}schel, S.~Riedel, P.~Lewis, A.~Bakhtin, Y.~Wu,
  A.~Miller, Language models as knowledge bases?, in: Proceedings of the 2019
  Conference on Empirical Methods in Natural Language Processing and the 9th
  International Joint Conference on Natural Language Processing (EMNLP-IJCNLP),
  Association for Computational Linguistics, 2019, pp. 2463--2473.
\newblock \href {http://dx.doi.org/10.18653/v1/D19-1250}
  {\path{doi:10.18653/v1/D19-1250}}.

\bibitem{shwartz2020unsupervised}
V.~Shwartz, P.~West, R.~Le~Bras, C.~Bhagavatula, Y.~Choi, Unsupervised
  commonsense question answering with self-talk, in: Proceedings of the 2020
  Conference on Empirical Methods in Natural Language Processing (EMNLP),
  Association for Computational Linguistics, 2020, pp. 4615--4629.
\newblock \href {http://dx.doi.org/10.18653/v1/2020.emnlp-main.373}
  {\path{doi:10.18653/v1/2020.emnlp-main.373}}.

\bibitem{do2021rotten}
N.~Do, E.~Pavlick, Are rotten apples edible? challenging commonsense inference
  ability with exceptions, in: Findings of the Association for Computational
  Linguistics: ACL-IJCNLP 2021, Association for Computational Linguistics,
  2021, pp. 2061--2073,
  \url{https://aclanthology.org/2021.findings-acl.181.pdf}.

\bibitem{kassner2020negated}
N.~Kassner, H.~Sch{\"u}tze, Negated and misprimed probes for pretrained
  language models: Birds can talk, but cannot fly, in: Proceedings of the 58th
  Annual Meeting of the Association for Computational Linguistics, Association
  for Computational Linguistics, 2020, pp. 7811--7818,
  \url{https://aclanthology.org/2020.acl-main.698.pdf?ref=https://githubhelp.com}.

\bibitem{tu2022misc}
Q.~Tu, Y.~Li, J.~Cui, B.~Wang, J.-R. Wen, R.~Yan, Misc: A mixed strategy-aware
  model integrating comet for emotional support conversation, in: Proceedings
  of the 60th Annual Meeting of the Association for Computational Linguistics
  (Volume 1: Long Papers), 2022, pp. 308--319.
\newblock \href {http://dx.doi.org/10.18653/v1/2022.acl-long.25}
  {\path{doi:10.18653/v1/2022.acl-long.25}}.

\bibitem{young2018augmenting}
T.~Young, E.~Cambria, I.~Chaturvedi, H.~Zhou, S.~Biswas, M.~Huang, Augmenting
  end-to-end dialogue systems with commonsense knowledge, in: Proceedings of
  the AAAI Conference on Artificial Intelligence, Vol.~32, 2018,
  \url{https://ojs.aaai.org/index.php/AAAI/article/view/11923}.

\bibitem{zhou2018commonsense}
H.~Zhou, T.~Young, M.~Huang, H.~Zhao, J.~Xu, X.~Zhu, Commonsense knowledge
  aware conversation generation with graph attention., in: IJCAI, 2018, pp.
  4623--4629, \url{https://www.ijcai.org/Proceedings/2018/0643.pdf}.

\bibitem{mihaylov2018knowledgeable}
T.~Mihaylov, A.~Frank, Knowledgeable reader: Enhancing cloze-style reading
  comprehension with external commonsense knowledge, in: Proceedings of the
  56th Annual Meeting of the Association for Computational Linguistics (Volume
  1: Long Papers), 2018, pp. 821--832.
\newblock \href {http://dx.doi.org/10.18653/v1/P18-1076}
  {\path{doi:10.18653/v1/P18-1076}}.

\bibitem{lal2022analyzing}
Y.~K. Lal, H.~Liu, N.~Tandon, N.~Chambers, R.~Mooney, N.~Balasubramanian,
  Analyzing the contribution of commonsense knowledge sources for why-question
  answering, in: ACL 2022 Workshop on Commonsense Representation and Reasoning,
  2022, \url{https://openreview.net/pdf?id=H4xz8zteub9}.

\bibitem{chen2019incorporating}
J.~Chen, J.~Chen, Z.~Yu, Incorporating structured commonsense knowledge in
  story completion, in: Proceedings of the AAAI Conference on Artificial
  Intelligence, Vol.~33, 2019, pp. 6244--6251,
  \url{https://ojs.aaai.org/index.php/AAAI/article/view/5183}.

\bibitem{wang2013common}
Q.-F. Wang, E.~Cambria, C.-L. Liu, A.~Hussain, Common sense knowledge for
  handwritten chinese text recognition, Cognitive Computation 5~(2) (2013)
  234--242.
\newblock \href {http://dx.doi.org/10.1007/s12559-012-9183-y}
  {\path{doi:10.1007/s12559-012-9183-y}}.

\bibitem{zhong2021care}
P.~Zhong, D.~Wang, P.~Li, C.~Zhang, H.~Wang, C.~Miao, Care: Commonsense-aware
  emotional response generation with latent concepts, in: Proceedings of the
  AAAI Conference on Artificial Intelligence, Vol.~35, 2021, pp. 14577--14585.
\newblock \href
  {http://dx.doi.org/https://ojs.aaai.org/index.php/AAAI/article/view/17713}
  {\path{doi:https://ojs.aaai.org/index.php/AAAI/article/view/17713}}.

\bibitem{zhu2021topic}
L.~Zhu, G.~Pergola, L.~Gui, D.~Zhou, Y.~He, Topic-driven and knowledge-aware
  transformer for dialogue emotion detection, in: Proceedings of the 59th
  Annual Meeting of the Association for Computational Linguistics and the 11th
  International Joint Conference on Natural Language Processing (Volume 1: Long
  Papers), 2021, pp. 1571--1582.
\newblock \href {http://dx.doi.org/10.18653/v1/2021.acl-long.125}
  {\path{doi:10.18653/v1/2021.acl-long.125}}.

\bibitem{speer2017conceptnet}
R.~Speer, J.~Chin, C.~Havasi, Conceptnet 5.5: An open multilingual graph of
  general knowledge, in: Proceedings of the AAAI Conference on Artificial
  Intelligence, Vol.~31, 2017,
  \url{https://dl.acm.org/doi/abs/10.5555/3298023.3298212}.

\bibitem{cambria2018senticnet}
E.~Cambria, S.~Poria, D.~Hazarika, K.~Kwok, Senticnet 5: Discovering conceptual
  primitives for sentiment analysis by means of context embeddings, in:
  Proceedings of the AAAI conference on artificial intelligence, Vol.~32, 2018,
  \url{https://ojs.aaai.org/index.php/AAAI/article/view/11559}.

\bibitem{mostafazadeh2020glucose}
N.~Mostafazadeh, A.~Kalyanpur, L.~Moon, D.~Buchanan, L.~Berkowitz, O.~Biran,
  J.~Chu-Carroll, Glucose: Generalized and contextualized story explanations,
  in: Proceedings of the 2020 Conference on Empirical Methods in Natural
  Language Processing (EMNLP), Association for Computational Linguistics, 2020,
  pp. 4569--4586.
\newblock \href {http://dx.doi.org/10.18653/v1/2020.emnlp-main.370}
  {\path{doi:10.18653/v1/2020.emnlp-main.370}}.

\bibitem{rashkin2018event2mind}
H.~Rashkin, M.~Sap, E.~Allaway, N.~A. Smith, Y.~Choi, Event2mind: Commonsense
  inference on events, intents, and reactions, in: Proceedings of the 56th
  Annual Meeting of the Association for Computational Linguistics (Volume 1:
  Long Papers), Association for Computational Linguistics, 2018, pp. 463--473.
\newblock \href {http://dx.doi.org/10.18653/v1/P18-1043}
  {\path{doi:10.18653/v1/P18-1043}}.

\bibitem{romero2020inside}
J.~Romero, S.~Razniewski, Inside quasimodo: Exploring construction and usage of
  commonsense knowledge, in: Proceedings of the 29th ACM International
  Conference on Information \& Knowledge Management, 2020, pp. 3445--3448.
\newblock \href {http://dx.doi.org/10.1145/3340531.3417416}
  {\path{doi:10.1145/3340531.3417416}}.

\bibitem{tandon2017webchild}
N.~Tandon, G.~De~Melo, G.~Weikum, Webchild 2.0: Fine-grained commonsense
  knowledge distillation, in: Proceedings of ACL 2017, System Demonstrations,
  Association for Computational Linguistics, 2017, pp. 115--120.
\newblock \href {http://dx.doi.org/10.18653/v1/P17-4020}
  {\path{doi:10.18653/v1/P17-4020}}.

\bibitem{gabriel2021paragraph}
S.~Gabriel, C.~Bhagavatula, V.~Shwartz, R.~Le~Bras, M.~Forbes, Y.~Choi,
  Paragraph-level commonsense transformers with recurrent memory, in:
  Proceedings of the AAAI Conference on Artificial Intelligence, Vol.~35, 2021,
  pp. 12857--12865,
  \url{https://ojs.aaai.org/index.php/AAAI/article/view/17521}.

\bibitem{kentonbert}
J.~D. M.-W.~C. Kenton, L.~K. Toutanova, Bert: Pre-training of deep
  bidirectional transformers for language understanding, in: Proceedings of the
  2019 Conference of the North {A}merican Chapter of the Association for
  Computational Linguistics: Human Language Technologies, Vol.~1, Association
  for Computational Linguistics, 2019, pp. 4171--4186.
\newblock \href {http://dx.doi.org/10.18653/v1/N19-1423}
  {\path{doi:10.18653/v1/N19-1423}}.

\bibitem{clark2020electra}
K.~Clark, M.-T. Luong, Q.~V. Le, C.~D. Manning, Electra: Pre-training text
  encoders as discriminators rather than generators, arXiv preprint
  arXiv:2003.10555\href {http://dx.doi.org/10.48550/arXiv.2003.10555}
  {\path{doi:10.48550/arXiv.2003.10555}}.

\bibitem{yang2019xlnet}
Z.~Yang, Z.~Dai, Y.~Yang, J.~Carbonell, R.~R. Salakhutdinov, Q.~V. Le, Xlnet:
  Generalized autoregressive pretraining for language understanding, Advances
  in neural information processing systems 32,
  \url{http://papers.neurips.cc/paper/8812-xlnet-generalized-autoregressive-pretraining-for-language-understanding.pdf}.

\bibitem{lan2019albert}
Z.~Lan, M.~Chen, S.~Goodman, K.~Gimpel, P.~Sharma, R.~Soricut, Albert: A lite
  bert for self-supervised learning of language representations, arXiv preprint
  arXiv:1909.11942\href {http://dx.doi.org/10.48550/arXiv.1909.11942}
  {\path{doi:10.48550/arXiv.1909.11942}}.

\bibitem{peters2018deep}
M.~E. Peters, M.~Neumann, M.~Iyyer, M.~Gardner, C.~Clark, K.~Lee,
  L.~Zettlemoyer, Deep contextualized word representations, in: Proceedings of
  the 2018 Conference of the North {A}merican Chapter of the Association for
  Computational Linguistics: Human Language Technologies,(Long Papers), Vol.~1,
  Association for Computational Linguistics, 2018, pp. 2227--2237.
\newblock \href {http://dx.doi.org/10.18653/v1/N18-1202}
  {\path{doi:10.18653/v1/N18-1202}}.

\bibitem{vaswani2017attention}
A.~Vaswani, N.~Shazeer, N.~Parmar, J.~Uszkoreit, L.~Jones, A.~N. Gomez,
  {\L}.~Kaiser, I.~Polosukhin, Attention is all you need, Advances in neural
  information processing systems 30 (2017) 6000–6010,
  \url{https://proceedings.neurips.cc/paper/2017/file/3f5ee243547dee91fbd053c1c4a845aa-Paper.pdf}.

\bibitem{lin2021survey}
T.~Lin, Y.~Wang, X.~Liu, X.~Qiu, A survey of transformers, arXiv preprint
  arXiv:2106.04554\href {http://dx.doi.org/10.48550/arXiv.2106.04554}
  {\path{doi:10.48550/arXiv.2106.04554}}.

\bibitem{gain2021iitp}
B.~Gain, D.~Bandyopadhyay, T.~Saikh, A.~Ekbal, Iitp@ coliee 2019: legal
  information retrieval using bm25 and bert, arXiv preprint
  arXiv:2104.08653\href {http://dx.doi.org/10.48550/arXiv.2104.08653}
  {\path{doi:10.48550/arXiv.2104.08653}}.

\bibitem{mishra2022please}
K.~Mishra, M.~Firdaus, A.~Ekbal, Please be polite: Towards building a
  politeness adaptive dialogue system for goal-oriented conversations,
  Neurocomputing 494 (2022) 242--254.
\newblock \href {http://dx.doi.org/10.1016/j.neucom.2022.04.029}
  {\path{doi:10.1016/j.neucom.2022.04.029}}.

\bibitem{yadav2021nlm}
S.~Yadav, M.~Sarrouti, D.~Gupta, Nlm at mediqa 2021: Transfer learning-based
  approaches for consumer question and multi-answer summarization, in:
  Proceedings of the 20th Workshop on Biomedical Language Processing, 2021, pp.
  291--301.
\newblock \href {http://dx.doi.org/10.18653/v1/2021.bionlp-1.34}
  {\path{doi:10.18653/v1/2021.bionlp-1.34}}.

\bibitem{yadav2022question}
S.~Yadav, D.~Gupta, A.~B. Abacha, D.~Demner-Fushman, Question-aware transformer
  models for consumer health question summarization, Journal of Biomedical
  Informatics 128 (2022) 104040.
\newblock \href {http://dx.doi.org/10.1016/j.jbi.2022.104040}
  {\path{doi:10.1016/j.jbi.2022.104040}}.

\bibitem{pingali2021multimodal}
S.~Pingali, S.~Yadav, P.~Dutta, S.~Saha, Multimodal graph-based transformer
  framework for biomedical relation extraction, in: Findings of the Association
  for Computational Linguistics: ACL-IJCNLP 2021, 2021, pp. 3741--3747,
  \url{https://aclanthology.org/2021.findings-acl.328.pdf}.

\bibitem{shin2019effective}
J.~Shin, Y.~Lee, K.~Jung, Effective sentence scoring method using bert for
  speech recognition, in: Asian Conference on Machine Learning, PMLR, 2019, pp.
  1081--1093, \url{https://proceedings.mlr.press/v101/shin19a.html}.

\bibitem{singh2022unity}
G.~V. Singh, M.~Firdaus, A.~Ekbal, P.~Bhattacharyya, Unity in diversity:
  Multilabel emoji identification in tweets, IEEE Transactions on Computational
  Social Systems\href {http://dx.doi.org/10.1109/TCSS.2022.3162865}
  {\path{doi:10.1109/TCSS.2022.3162865}}.

\bibitem{conneau2020unsupervised}
A.~Conneau, K.~Khandelwal, N.~Goyal, V.~Chaudhary, G.~Wenzek, F.~Guzm{\'a}n,
  {\'E}.~Grave, M.~Ott, L.~Zettlemoyer, V.~Stoyanov, Unsupervised cross-lingual
  representation learning at scale, in: Proceedings of the 58th Annual Meeting
  of the Association for Computational Linguistics, Association for
  Computational Linguistics, 2020, pp. 8440--8451.
\newblock \href {http://dx.doi.org/10.18653/v1/2020.acl-main.747}
  {\path{doi:10.18653/v1/2020.acl-main.747}}.

\bibitem{liu2020multilingual}
Y.~Liu, J.~Gu, N.~Goyal, X.~Li, S.~Edunov, M.~Ghazvininejad, M.~Lewis,
  L.~Zettlemoyer,
  \href{https://direct.mit.edu/tacl/article/doi/10.1162/tacl_a_00343/96484/Multilingual-Denoising-Pre-training-for-Neural}{Multilingual
  denoising pre-training for neural machine translation}, Transactions of the
  Association for Computational Linguistics 8 (2020) 726--742.
\newblock \href {http://dx.doi.org/10.1162/tacl_a_00343}
  {\path{doi:10.1162/tacl_a_00343}}.
\newline\urlprefix\url{https://direct.mit.edu/tacl/article/doi/10.1162/tacl_a_00343/96484/Multilingual-Denoising-Pre-training-for-Neural}

\bibitem{kumar2021sentiment}
A.~Kumar, V.~H.~C. Albuquerque, Sentiment analysis using xlm-r transformer and
  zero-shot transfer learning on resource-poor indian language, Transactions on
  Asian and Low-Resource Language Information Processing 20~(5) (2021) 1--13.
\newblock \href {http://dx.doi.org/10.1145/3461764}
  {\path{doi:10.1145/3461764}}.

\bibitem{novak2022document}
E.~Novak, L.~Bizjak, D.~Mladeni{\'c}, M.~Grobelnik, Why is a document relevant?
  understanding the relevance scores in cross-lingual document retrieval,
  Knowledge-Based Systems 244 (2022) 108545.
\newblock \href {http://dx.doi.org/10.1016/j.knosys.2022.108545}
  {\path{doi:10.1016/j.knosys.2022.108545}}.

\bibitem{muller2021first}
B.~Muller, Y.~Elazar, B.~Sagot, D.~Seddah, First align, then predict:
  Understanding the cross-lingual ability of multilingual bert, in: Proceedings
  of the 16th Conference of the European Chapter of the Association for
  Computational Linguistics: Main Volume, Association for Computational
  Linguistics, 2021, pp. 2214--2231.
\newblock \href {http://dx.doi.org/10.18653/v1/2021.eacl-main.189}
  {\path{doi:10.18653/v1/2021.eacl-main.189}}.

\bibitem{cer2018universal}
D.~Cer, Y.~Yang, S.-y. Kong, N.~Hua, N.~Limtiaco, R.~S. John, N.~Constant,
  M.~Guajardo-Cespedes, S.~Yuan, C.~Tar, et~al., Universal sentence encoder,
  arXiv preprint arXiv:1803.11175\href
  {http://dx.doi.org/10.48550/arXiv.1803.11175}
  {\path{doi:10.48550/arXiv.1803.11175}}.

\bibitem{reimers2020making}
N.~Reimers, I.~Gurevych, Making monolingual sentence embeddings multilingual
  using knowledge distillation, in: Proceedings of the 2020 Conference on
  Empirical Methods in Natural Language Processing (EMNLP), Association for
  Computational Linguistics, 2020, pp. 4512--4525.
\newblock \href {http://dx.doi.org/10.18653/v1/2020.emnlp-main.365}
  {\path{doi:10.18653/v1/2020.emnlp-main.365}}.

\bibitem{feng2022language}
F.~Feng, Y.~Yang, D.~Cer, N.~Arivazhagan, W.~Wang, Language-agnostic bert
  sentence embedding, in: Proceedings of the 60th Annual Meeting of the
  Association for Computational Linguistics (Volume 1: Long Papers),
  Association for Computational Linguistics, 2022, pp. 878--891.
\newblock \href {http://dx.doi.org/10.18653/v1/2022.acl-long.62}
  {\path{doi:10.18653/v1/2022.acl-long.62}}.

\bibitem{chaudhary2019low}
V.~Chaudhary, Y.~Tang, F.~Guzm{\'a}n, H.~Schwenk, P.~Koehn, Low-resource corpus
  filtering using multilingual sentence embeddings, in: Proceedings of the
  Fourth Conference on Machine Translation (Volume 3: Shared Task Papers, Day
  2), Association for Computational Linguistics, 2019, pp. 261--266.
\newblock \href {http://dx.doi.org/10.18653/v1/W19-5435}
  {\path{doi:10.18653/v1/W19-5435}}.

\bibitem{mohammad2021gated}
A.-S. Mohammad, M.~M. Hammad, A.~Sa’ad, A.-T. Saja, E.~Cambria, Gated
  recurrent unit with multilingual universal sentence encoder for arabic
  aspect-based sentiment analysis, Knowledge-Based Systems (2021) 107540\href
  {http://dx.doi.org/10.1016/j.knosys.2021.107540}
  {\path{doi:10.1016/j.knosys.2021.107540}}.

\bibitem{conneau2017supervised}
A.~Conneau, D.~Kiela, H.~Schwenk, L.~Barrault, A.~Bordes, Supervised learning
  of universal sentence representations from natural language inference data,
  in: Proceedings of the 2017 Conference on Empirical Methods in Natural
  Language Processing, Association for Computational Linguistics, 2017, pp.
  670--680.
\newblock \href {http://dx.doi.org/10.18653/v1/D17-1070}
  {\path{doi:10.18653/v1/D17-1070}}.

\bibitem{MBFC}
MBFC, Media bias/fact check - search and learn the bias of news media,
  \url{https://mediabiasfactcheck.com}, accessed: June 6, 2022.

\bibitem{baly2018predicting}
R.~Baly, G.~Karadzhov, D.~Alexandrov, J.~Glass, P.~Nakov, Predicting factuality
  of reporting and bias of news media sources, in: Proceedings of the 2018
  Conference on Empirical Methods in Natural Language Processing, Association
  for Computational Linguistics, 2018, pp. 3528--3539.
\newblock \href {http://dx.doi.org/10.18653/v1/D18-1389}
  {\path{doi:10.18653/v1/D18-1389}}.

\bibitem{resnick2018iffy}
P.~Resnick, A.~Ovadya, G.~Gilchrist, Iffy quotient: A platform health metric
  for misinformation, Center for Social Media Responsibility 17,
  \url{http://umsi.info/iffy-quotient-whitepaper}.

\bibitem{MBFCMethodologyBIAS}
MBFC, Left vs. right bias: How we rate the bias of media sources - media
  bias/fact check,
  \url{https://mediabiasfactcheck.com/left-vs-right-bias-how-we-rate-the-bias-of-media-sources/},
  accessed: June 6, 2022.

\bibitem{MBFCMethodology}
MBFC, methodology - media bias/fact check,
  \url{https://mediabiasfactcheck.com/methodology/}, accessed: June 6, 2022.

\bibitem{LebanGregorEventRegistry}
G.~Leban, B.~Fortuna, J.~Brank, M.~Grobelnik, Event registry: learning about
  world events from news, in: Proceedings of the 23rd International Conference
  on World Wide Web, 2014, pp. 107--110.
\newblock \href {http://dx.doi.org/10.1145/2567948.257702}
  {\path{doi:10.1145/2567948.257702}}.

\bibitem{swati2021eveout}
S.~Swati, D.~Mladeni{\'c}, T.~Erjavec, Eveout: an event-centric news dataset to
  analyze an outlet's event selection patterns, Informatica 45~(7).
\newblock \href {http://dx.doi.org/10.31449/inf.v45i7.3410}
  {\path{doi:10.31449/inf.v45i7.3410}}.

\bibitem{mladenic2020you}
S.~Swati, D.~Mladeni{\'c}, Are you following the right news-outlet? a machine
  learning based approach to outlet prediction, in: In Proceedings of the
  Slovenian KDD Conference on Data Mining and Data Warehouses (SiKDD), 2020,
  \url{https://ailab.ijs.si/Dunja/SiKDD2020/Papers/08 -
  swati_outlet_prediction.pdf}.

\bibitem{mladenic2021understanding}
S.~Swati, D.~Mladeni{\'c}, Understanding the impact of geographical bias on
  news sentiment: A case study on london and rio olympics, in: In Proceedings
  of the Slovenian KDD Conference on Data Mining and Data Warehouses (SiKDD),
  2021, \url{https://ailab.ijs.si/dunja/SiKDD2021/Papers/Swati+Mladenic.pdf}.

\bibitem{bosselut2019comet}
A.~Bosselut, H.~Rashkin, M.~Sap, C.~Malaviya, A.~Celikyilmaz, Y.~Choi, Comet:
  Commonsense transformers for automatic knowledge graph construction, in:
  Proceedings of the 57th Annual Meeting of the Association for Computational
  Linguistics, Association for Computational Linguistics, 2019, pp. 4762--4779.
\newblock \href {http://dx.doi.org/10.18653/v1/P19-1470}
  {\path{doi:10.18653/v1/P19-1470}}.

\bibitem{nantomah2019some}
K.~Nantomah, On some properties of the sigmoid function, Asia
  Mathematika\url{https://hal.archives-ouvertes.fr/hal-02635089/}.

\bibitem{majumder2020mime}
N.~Majumder, P.~Hong, S.~Peng, J.~Lu, D.~Ghosal, A.~Gelbukh, R.~Mihalcea,
  S.~Poria, Mime: Mimicking emotions for empathetic response generation, in:
  Proceedings of the 2020 Conference on Empirical Methods in Natural Language
  Processing (EMNLP), Association for Computational Linguistics, 2020, pp.
  8968--8979.
\newblock \href {http://dx.doi.org/10.18653/v1/2020.emnlp-main.721}
  {\path{doi:10.18653/v1/2020.emnlp-main.721}}.

\bibitem{kingma2014adam}
D.~P. Kingma, J.~Ba, Adam: A method for stochastic optimization, in:
  International Conference on Learning Representations (ICLR), 2015, pp. 1--13,
  \url{https://hdl.handle.net/11245/1.505367}.

\bibitem{sayar2020leveraging}
U.~N. Sayar Ghosh~Roy, T.~Raha, Z.~Abid, V.~Varma, Leveraging multilingual
  transformers for hate speech detection, CEUR Workshop Proceedings: FIRE 2020
  - Forum for Information Retrieval
  Evaluation\url{http://ceur-ws.org/Vol-2826/T2-4.pdf}.

\bibitem{yang2020multilingual}
Y.~Yang, D.~Cer, A.~Ahmad, M.~Guo, J.~Law, N.~Constant, G.~H. Abrego, S.~Yuan,
  C.~Tar, Y.-H. Sung, et~al., Multilingual universal sentence encoder for
  semantic retrieval, in: Proceedings of the 58th Annual Meeting of the
  Association for Computational Linguistics: System Demonstrations, 2020, pp.
  87--94.
\newblock \href {http://dx.doi.org/10.18653/v1/2020.acl-demos.12}
  {\path{doi:10.18653/v1/2020.acl-demos.12}}.

\bibitem{devlin2018bert}
J.~Devlin, M.-W. Chang, K.~Lee, K.~Toutanova, Bert: Pre-training of deep
  bidirectional transformers for language understanding, arXiv preprint
  arXiv:1810.04805\href {http://dx.doi.org/10.48550/arXiv.1810.04805}
  {\path{doi:10.48550/arXiv.1810.04805}}.

\bibitem{kowsari2019text}
K.~Kowsari, K.~Jafari~Meimandi, M.~Heidarysafa, S.~Mendu, L.~Barnes, D.~Brown,
  Text classification algorithms: A survey, Information 10~(4) (2019) 150.
\newblock \href {http://dx.doi.org/10.3390/info10040150}
  {\path{doi:10.3390/info10040150}}.

\bibitem{real1996probabilistic}
R.~Real, J.~M. Vargas, The probabilistic basis of jaccard's index of
  similarity, Systematic biology 45~(3) (1996) 380--385.
\newblock \href {http://dx.doi.org/10.2307/2413572}
  {\path{doi:10.2307/2413572}}.

\bibitem{nagle1998proposal}
B.~Nagle, A proposal for dealing with grade inflation: The relative performance
  index, Journal of Education for Business 74~(1) (1998) 40--43.
\newblock \href {http://dx.doi.org/10.1080/08832329809601659}
  {\path{doi:10.1080/08832329809601659}}.

\end{thebibliography}

\end{document}